\documentclass{article}

\usepackage{microtype}
\usepackage{graphicx}
\usepackage{subcaption}
\usepackage{booktabs} 

\usepackage{hyperref}

\usepackage[accepted]{icml2026}

\usepackage{amsmath}
\usepackage{amssymb}
\usepackage{mathtools}
\usepackage{amsthm}
\usepackage{amsfonts}
\usepackage{array}
\usepackage{booktabs}
\usepackage{multirow}
\usepackage{xcolor}
\usepackage[table]{xcolor}

\usepackage[capitalize,noabbrev]{cleveref}

\theoremstyle{plain}

\theoremstyle{definition}

\theoremstyle{remark}

\icmltitlerunning{Large Language Model Teaches Visual Students: Cross-Modality Transfer of Fine-Grained Conceptual Knowledge}

\newcommand{\LaViD}{\text{LaViD}}

\begin{document}

\twocolumn[
  \icmltitle{Large Language Model Teaches Visual Students: Cross-Modality Transfer of Fine-Grained Conceptual Knowledge}

  \icmlsetsymbol{equal}{*}

  \begin{icmlauthorlist}
    \icmlauthor{Thomas Shih-Chao Liang}{equal,yyy}
    \icmlauthor{Zhuoran Yu}{equal,yyy}
    \icmlauthor{Yong Jae Lee}{yyy}
  \end{icmlauthorlist}

  \icmlaffiliation{yyy}{University of Wisconsin-Madison}

  \icmlcorrespondingauthor{Thomas Shih-Chao Liang}{liangt@cs.wisc.edu}

  \icmlkeywords{Machine Learning, ICML, Knowledge Distillation, Cross-Modal Learning, Fine-Grained Visual Classification, Large Language Models}

  \vskip 0.3in
]

\printAffiliationsAndNotice{\icmlEqualContribution}

\begin{abstract}
Large Language Models (LLMs) possess broad conceptual knowledge acquired through large-scale text pretraining, yet their potential to supervise models in other modalities remains underexplored. In this work, we propose \LaViD—Language-to-Visual Knowledge Distillation—a simple and effective framework for transferring high-level semantic knowledge from a language-only teacher to a vision-only student model. Instead of relying on paired multimodal data, \LaViD{} elicits conceptual signals from an LLM by prompting it to generate multiple-choice questions (MCQs) that probe semantic distinctions between visual classes. Each class is mapped to a soft label distribution over these MCQs, forming a rich conceptual signature that guides the student through an auxiliary distillation loss. Notably, despite using a language-only teacher without access to image data, \LaViD{} consistently outperforms recent methods like MaKD that distill from vision-language models across multiple fine-grained benchmarks. It also achieves competitive or superior performance compared to state-of-the-art visual distillation methods such as DKD and MLKD, with further gains when combined with logit standardization. On the Waterbirds dataset, \LaViD{} substantially improves worst-group accuracy, demonstrating enhanced robustness to spurious correlations with distillation. Code is available at \url{https://github.com/lliangthomas/lavid}.
\end{abstract}

\section{Introduction}
\label{sec:intro}

Knowledge Distillation (KD)~\cite{buciluǎ2006model, hinton2015distilling, romero2014fitnets} is a foundational technique to transfer knowledge from a large teacher model to a smaller student model. This approach~\cite{tian2019contrastive, chen2022knowledge, yang2021knowledge, zhao2022decoupled, hao2023one} typically requires a dataset-specific teacher that guides the student through the learning process through logits~\cite{hinton2015distilling, tian2019contrastive, hao2023one, sun2024logit, zhao2022decoupled} or feature representations of the teacher~\cite{romero2014fitnets, chen2022knowledge, zhang2020improve, park2019relational, tung2019similarity}. However, this reliance on purely visual supervision is often insufficient for Fine-Grained Visual Classification, where models must discern subtle inter-class distinctions and overfit to spurious background correlations instead of learning robust traits.

The development of Large Language Models (LLMs)~\cite{touvron2023llama, grattafiori2024llama, touvron2023llama2, chowdhery2023palm, brown2020language, yang2024qwen2, raffel2020exploring} has revolutionized the field. What’s particularly interesting about the knowledge encoded in a large language model is its \textit{conceptual} nature: it often transcends the textual modality it is trained on. For example, describing a ``cat'' in language---``a small, furry animal with pointed ears and whiskers''---carries the same conceptual meaning as visually identifying a cat in an image. While the modality of the input differs (text vs.\ image), the underlying notion of ``catness'' remains the same. This observation suggests that the knowledge stored in LLMs is not tied to language, but instead reflects general, modality-agnostic concepts, as hypothesized in~\cite{huh2024platonic}. It raises a compelling question: 

\textit{Can such conceptual knowledge, encoded purely in text, be transferred to guide visual learning?}

In this work, we introduce \textbf{L}anguage-to-\textbf{Vi}sual Knowledge \textbf{D}istillation (\LaViD{}), a simple yet effective approach that distills general knowledge from text-only large language models (LLMs) into visual student models. Rather than relying on paired multimodal data or task-specific supervision, \LaViD{} uses the broad world knowledge encoded in LLMs to provide conceptual guidance. It does so by eliciting structured and interpretable signals through multiple-choice questions that probe semantic distinctions between classes. This allows visual models to learn not just from labeled data, but also from external textual knowledge—bridging the gap between language and vision without requiring aligned inputs.

\LaViD{} distills conceptual knowledge from a language-only teacher into a visual student model using a two-stage process. First, we prompt an LLM with dataset metadata to generate multiple-choice questions (MCQs) that probe semantic differences between classes. Each question includes a placeholder token (\texttt{<object>}), which is replaced with each class name to instantiate class-specific prompts. The LLM’s pre-softmax logits over answer options are extracted and normalized into soft label distributions, forming a semantic signature for each class. Next, the student processes input images and predicts auxiliary logits aligned with the LLM’s question space. Training minimizes a standard classification loss along with a mean squared error (MSE) loss between the student’s auxiliary predictions and the LLM-derived targets. 

The core intuition behind \LaViD{} is that LLMs encode structured world knowledge, enabling them to express nuanced conceptual relationships between categories. By prompting the LLM with class-specific multiple-choice questions, we elicit semantic distinctions—such as coloration, shape, or behavior—that define how different classes relate at a conceptual level. The resulting logits provide a structured view of inter-class similarities and differences, which we use as supervision to guide the student model. Unlike fixed class embeddings or similarity targets, these conceptual signatures are structured across multiple semantic dimensions induced by diverse questions, providing richer relational supervision than conventional label smoothing or representation matching. This signal pushes the student to organize its internal representations around meaningful attributes, promoting deeper generalization beyond rote memorization of class labels.

We evaluate \LaViD{} across six fine-grained classification benchmarks and find it consistently outperforms both traditional KD methods and multimodal LLM-based baselines. Notably, despite using a language-only teacher with \textbf{no access to image data}, \LaViD{} surpasses recent approaches MaKD~\cite{lee2025multi} that distill from vision-language models InternVL~\cite{chen2024internvl}. It also achieves competitive or superior performance compared to state-of-the-art KD methods like DKD~\cite{zhao2022decoupled} and MLKD~\cite{jin2023multi}, and can be further combined with logit standardization~\cite{sun2024logit} for additional gains. Beyond accuracy, we demonstrate that \LaViD{} mitigates dataset bias: on the Waterbirds dataset~\cite{sagawa2019distributionally}, it significantly improves worst-group accuracy, indicating improved robustness to spurious correlations. Extensive ablation studies further validate the importance of our design choices, including the use of MCQs, LLMs, and their semantic structure.

\section{Related Work}
\label{sec:related}

\textbf{Knowledge Distillation.} Knowledge distillation (KD) generally focuses on transferring knowledge from a larger teacher to a smaller student~\cite{buciluǎ2006model, hinton2015distilling}. In the unimodal setting, this process occurs within the same modality, where early work focused on matching logits~\cite{hinton2015distilling} or intermediate features~\cite{romero2014fitnets}. Later studies extended KD across heterogeneous architectures~\cite{liu2022cross, zhu2023good} and demonstrated greater gains with large teacher--student performance gaps~\cite{huang2022knowledge, fan2024scalekd}. However, conventional KD typically requires training a dataset-specific teacher, which adds computational overhead and risks transferring dataset biases~\cite{ojha2023knowledge}. Cross-modal knowledge distillation transfers supervision across different modalities~\cite{xue2021multimodal, xue2022modality, garcia2018modality, gupta2016cross}. This has supported semantic generalization in open-vocabulary recognition, where students align with textual embeddings or leverage CLIP's image encoder~\cite{radford2021learning, gu2021open, wu2023aligning, xu2023masqclip}. Unlike these approaches, which rely on paired inputs or multimodal encoders, \LaViD{} distills general world knowledge from a language-only teacher to a vision-only student \textit{without paired data, modality alignment, or a shared embedding space}.

\textbf{Fine-Grained Classification.} Fine-Grained Visual Classification focuses on distinguishing classes within a broader meta-class, often containing subtle and challenging inter-class differences. Typically, these approaches are separated into localization (explicit discriminative regions) ~\cite{ge2019weaklysupervisedcomplementaryparts, Wang_Wang_Li_Dou_Li_2020, schmidt2025saccadicvisionfinegrainedvisual} and feature-encoding (implicit or conceptual differences) ~\cite{lin2017bilinearcnnsfinegrainedvisual, zheng2019learningdeepbilineartransformation}. While DFKD-FGVC~\cite{shao2024datafreeknowledgedistillationfinegrained} recently proposed a KD method for fine-grained tasks, it remains constrained to homogeneous teacher--student architectures. Our work differentiates itself by demonstrating that a visual student can learn these nuanced distinctions directly from \textit{conceptual} knowledge, without requiring aligned modalities or a homogeneous teacher.
\section{Methodology}
\label{sec:method}

\begin{figure*}[h!]
    \centering
    \includegraphics[scale=0.6]{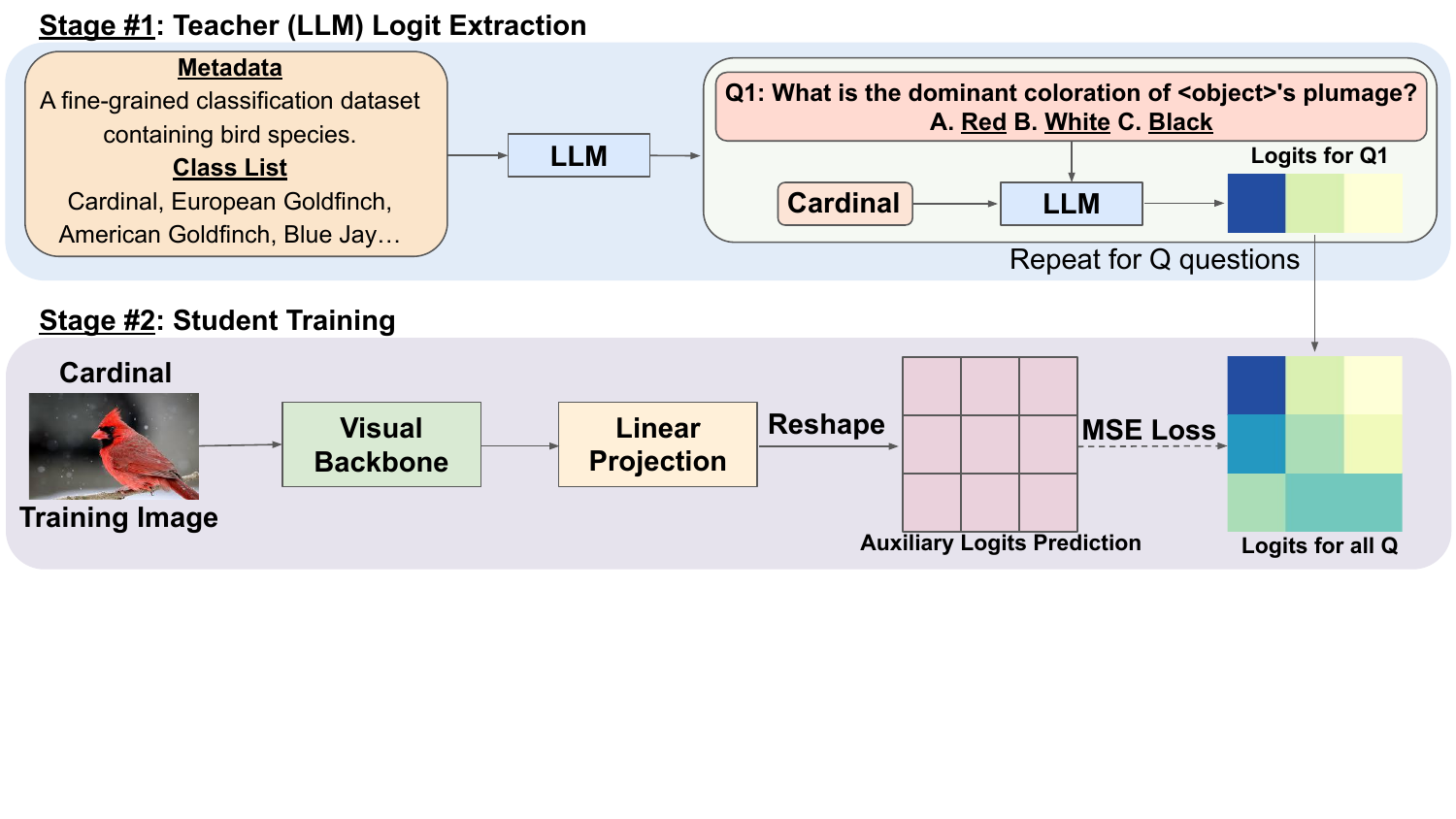}
    \vspace{-2cm}
    \caption{\textbf{Overview of \LaViD{}.} \textbf{Stage \#1}: The LLM is prompted with class metadata and names to generate diverse multiple-choice questions (MCQs) that capture high-level semantic differences. These are instantiated with each class to extract soft label distributions over answer options, forming a conceptual signature per class. \textbf{Stage \#2}: The student processes an image through a visual backbone and auxiliary head to predict logits aligned with the LLM’s question space. It is trained with a standard classification loss (not shown) and an auxiliary MSE loss against the LLM-derived targets.}
    \label{fig:method_pipeline}
\end{figure*}

In this section, we present \textbf{L}anguage-to-\textbf{Vi}sual Knowledge \textbf{D}istillation (\LaViD{}), a framework for transferring conceptual knowledge from a \textbf{text-only large language model (LLM)} to a purely \textbf{visual student model}. Unlike multimodal approaches that rely on paired vision–language inputs, \LaViD{} distills structured semantic supervision from language alone to guide visual representation learning. Concretely, \LaViD{} elicits relational conceptual knowledge from the LLM through structured semantic queries and aligns the student with the resulting multi-dimensional class relationships. While \LaViD{} adopts a distillation-style objective, it differs fundamentally from conventional knowledge distillation, which transfers instance-level predictions from task-trained teachers; instead, it reframes distillation as a mechanism for cross-modal concept transfer, injecting external world knowledge into visual learners rather than mimicking teacher outputs. Importantly, the conceptual targets in \LaViD{} are fixed per class but structured across diverse semantic dimensions induced by the generated questions, distinguishing them from class embeddings or label smoothing schemes that capture only coarse similarity structure. This structured semantic regularization encourages visual representations to align with meaningful conceptual factors, which we find contributes to improved robustness against spurious correlations in practice.

\subsection{Overview}

Let \( \mathcal{X} \) be the dataset, where \( \mathcal{X} = \{ \mathbf{x}_i \}_{i=1}^{N} \) and \( \mathbf{x}_i \in \mathbb{R}^{3 \times H \times W} \) represents the input image of size \( 3 \times H \times W \). Let \( \mathcal{Y} = \{ y_i \}_{i=1}^{N} \) be the corresponding set of labels, where \( y_i \in \{0,1\}^k \) is the one-hot encoded class label for the \( i \)-th sample, with \( k \) being the number of classes.

We define the total loss \( L \) as the sum of the supervised loss \( L_{\text{sup}} \) and the distillation loss \( L_{\text{LaViD}} \):

\[
L = L_{\text{sup}} + \lambda L_{\text{LaViD}}
\]

The supervised loss \( L_{\text{sup}} \) is the standard cross-entropy loss for classification:

\[
L_{\text{sup}} = - \frac{1}{N} \sum_{i=1}^{N} \sum_{c=1}^{k} y_{i,c} \log p(y_i = c | \mathbf{x}_i)
\]

where \( p(y_i = c | \mathbf{x}_i) \) is the predicted probability for class \( c \) for the \( i \)-th sample.

\begin{figure*}[h!]
    \centering
    \includegraphics[scale=0.6]{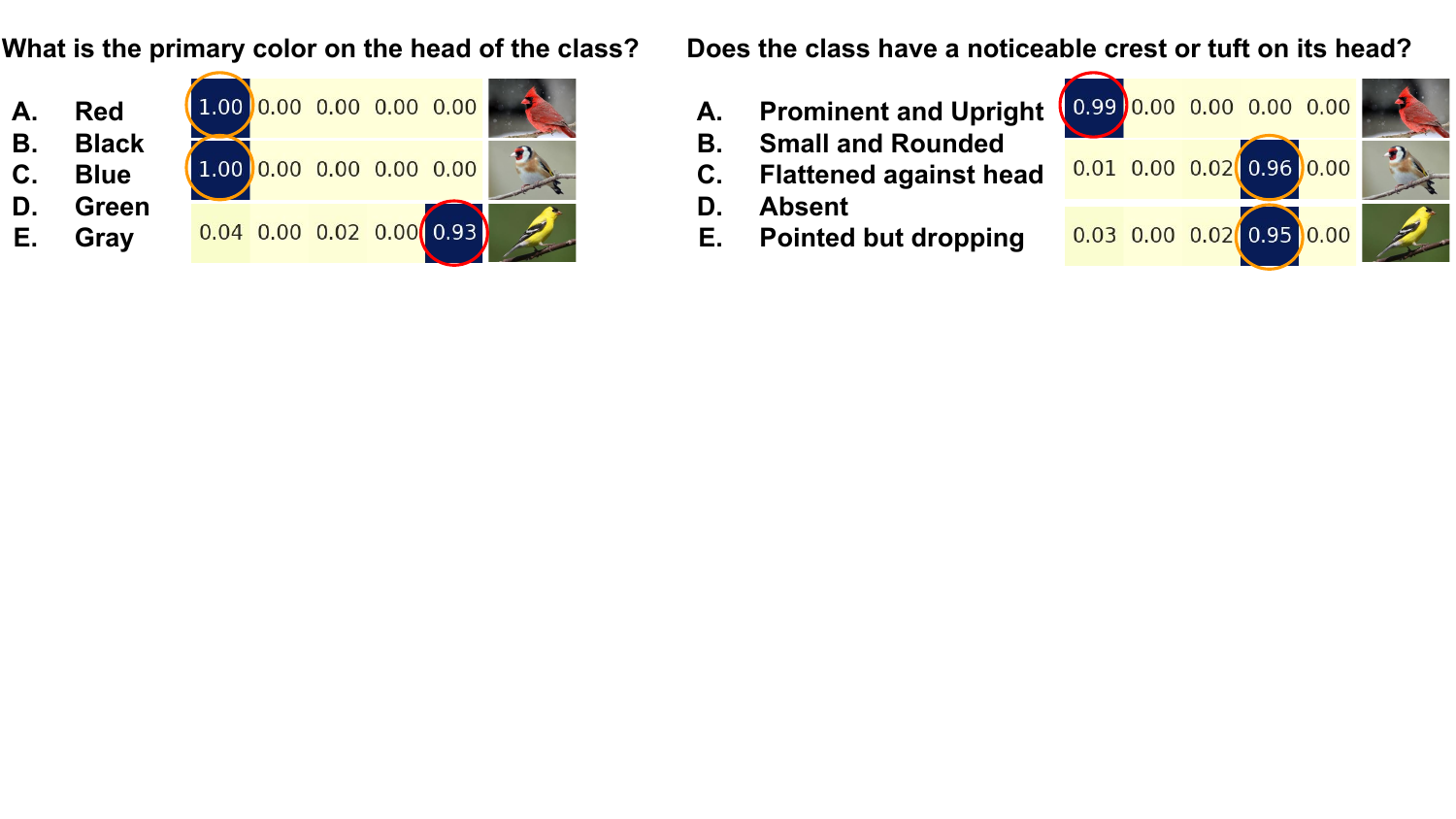}
    \vspace{-5.5cm}
    
\begin{tabular}{>{\centering\arraybackslash}m{0.25\linewidth}
                >{\centering\arraybackslash}m{0.25\linewidth}
                >{\centering\arraybackslash}m{0.25\linewidth}}
    \includegraphics[height=1.8em]{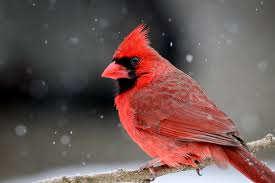} &
    \includegraphics[height=1.8em]{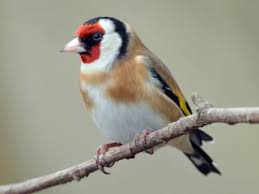} &
    \includegraphics[height=1.8em]{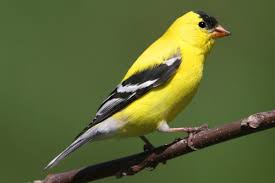} \\
    \small Cardinal & \small European Goldfinch & \small American Goldfinch
\end{tabular}

\caption{
\textbf{Structured semantics from LLM supervision.} The heatmap shows LLM logits for two questions where the \textit{European Goldfinch} aligns with the \textit{Cardinal} on head color (left) but with the \textit{American Goldfinch} on crest absence (right). These patterns provide relational supervision beyond standard class labels.
}
\label{fig:qualitative}
\end{figure*}

\subsection{Logit Extraction from the LLM}

To construct the distillation targets used in the \LaViD~loss, we first prompt the LLM to generate a set of multiple-choice questions that probe semantic distinctions between classes. These questions are then instantiated per class and used to query the LLM for logits over answer options. The resulting distributions serve as supervision signals to guide the student model during training. We emphasize that MCQ generation and logit extraction are performed once per dataset (rather than per training sample), making the supervision cost negligible relative to student training and eliminating the need to train or run large multimodal teachers during learning.

\textbf{Multiple-choice question generation.}
Let \(\mathcal{C} = \{c_1, \ldots, c_k\}\) denote the set of class names in the dataset. Given \(\mathcal{C}\) and dataset metadata, we prompt the LLM to generate a set of multiple-choice questions (MCQs) aimed at distinguishing between these classes. Each question must focus on a visually grounded concept, include a special \texttt{<object>} token as a placeholder for a target class, and assign each class to exactly one answer option. The answer choices must not include any class names or direct references as this would not force the LLM to think semantically. We use a single prompt to collect a set of \(Q\) such questions, each accompanied by \(M\) answer options. The full prompt is in the appendix.

\textbf{Per-class logit extraction.}
Once the MCQs are obtained, we instantiate each question by replacing the \texttt{<object>} token with the name of a specific class \(c \in \mathcal{C}\), resulting in a complete prompt. Each prompt is formatted using a chat interface, where the user poses the question followed by a list of labeled answer options (e.g., ``A. option one'', ``B. option two'', etc.), and the assistant is expected to reply with the correct option label (e.g., ``A''). We extract the pre-softmax logits for the next-token prediction following the assistant’s response prompt, focusing on the logits assigned to the first token of each answer label (``A'', ``B'', etc.). This process is repeated for all \(Q\) questions. For each class \(c\), the resulting LLM supervision takes the form of a \(Q \times M\) matrix, where each row corresponds to a question and contains a softmax-normalized probability distribution over the \(M\) options. These per-class matrices serve as the distillation targets for training the student model.

\subsection{Student Training with Distillation Loss}

To align the student model with the conceptual supervision from the LLM, we equip the visual backbone with an auxiliary linear head that maps the final feature vector into a flattened output of dimension \(QM\). Specifically, given an image \(x\), the student produces a feature vector \(f(x)\), which is projected to a vector \(s(x) \in \mathbb{R}^{QM}\). We reshape this into a \(Q \times M\) matrix, denoted \(S(x) \in \mathbb{R}^{Q \times M}\), which represents the student’s predictions over \(M\) answer options for each of the \(Q\) questions.

Each image is associated with a ground-truth class label \(y\), which indexes a class in \(\mathcal{C}\), and the corresponding LLM supervision matrix \(T_y \in \mathbb{R}^{Q \times M}\) serves as the soft target. The distillation loss is defined as:
\[
L_{\text{\LaViD}}(x, y) = \frac{1}{QM} \left\| S(x) - T_y \right\|_2^2.
\]

This loss guides the student to align with the class-level conceptual knowledge encoded by the LLM, providing an auxiliary training signal alongside conventional supervision.

\subsection{Structuring Class Semantics through Language}
\label{sec:qualitative}

To illustrate the intuition behind \LaViD{}, we present a toy example with two representative multiple-choice questions generated by GPT-4o and three bird species from the CUB dataset: \textit{Cardinal}, \textit{European Goldfinch}, and \textit{American Goldfinch}. Each question targets a visually grounded trait, such as head color or presence of a crest, and another language model produces logits over answer options for each class. These logits, shown in Figure~\ref{fig:qualitative}, reveal consistent and interpretable semantic structure.

Notably, the \textit{European Goldfinch} shares the same predicted head color as the \textit{Cardinal} (``Red'') but disagrees on the crest question, where the \textit{Cardinal} is ``Prominent and Upright'' while the goldfinch is ``Absent.'' Conversely, the \textit{European} and \textit{American Goldfinch} differ on head color but align on crest absence. These relationships are not incidental: they reflect consistent semantic distinctions captured by the LLM and transferred to the student model during distillation.

When the student is trained on a \textit{European Goldfinch} image, it is encouraged to produce auxiliary logits that agree with the \textit{Cardinal} on the head color question but diverge on the crest question. In contrast, training on the \textit{American Goldfinch} leads to agreement with the \textit{European Goldfinch} on crest absence but not head color. These supervision signals introduce structured relational constraints that reflect \textbf{external conceptual knowledge} captured by the LLM. \LaViD{} encourages the student to shape its internal representations in a way that reflects semantic and visual relationships across classes. 

These structured patterns encourage the student to embed visual classes in a space where both intra-class consistency and inter-class structure are preserved, aligned with the knowledge encoded in language. Conceptually, \LaViD{} acts as a semantic regularizer that biases visual representations toward world-knowledge-consistent class relationships, offering a potential explanation for its robustness benefits.

\section{Experiments}
\label{sec:experiments}
\begin{table*}[ht!]
\centering
\resizebox{0.9\linewidth}{!}{%
\begin{tabular}{lcccccccc}
\toprule
\textbf{Student} & \textbf{Teacher} & \textbf{Method} & \textbf{CUB} & \textbf{Caltech} & \textbf{Flowers} & \textbf{Aircraft} & \textbf{Pets} & \textbf{Cars} \\
\midrule
\multirow{11}{*}{RN-18} & -- & Ind Student & 63.07 & 78.65 & 75.73 & 78.95 & 77.57 & 85.77 \\
 & RN-50 & KD~\cite{hinton2015distilling}      &   64.76   &   80.07    &   76.37    &    81.16   &    79.32   &   86.94    \\
 & RN-50 & RKD~\cite{park2019relational}      &    61.70   &   78.45    &   73.82    &   78.18    &   78.91    &   85.78    \\
 & RN-50 & DKD~\cite{zhao2022decoupled}      &   67.67    &    80.02   &    76.35   &   84.48    &    82.23   &   89.31    \\
 & RN-50 & MLKD~\cite{jin2023multi}     &  68.76    &   80.66   &   76.58    &   85.35    &    82.94   &   90.23    \\
  & RN-50 & LS~\cite{sun2024logit}     &  \underline{70.42}     &   \underline{83.50}    &  \underline{82.34}     &  \underline{86.02}     &     \underline{83.56}  &    \underline{90.98}   \\
  
   & \cellcolor{gray!10}QN+R50 &  \cellcolor{gray!10}w/ \LaViD~(Ours)  &\cellcolor{gray!10}\textbf{72.46}  & \cellcolor{gray!10}\textbf{84.17}   &  \cellcolor{gray!10}\textbf{85.05}  & \cellcolor{gray!10}\textbf{86.21}  &  \cellcolor{gray!10}\textbf{85.09}& \cellcolor{gray!10}\textbf{91.20} \\
   \cmidrule(lr){2-9}
 & InternVL & MaKD~\cite{lee2025multi}     & 68.19 & 79.91 & 80.18 & 80.45 & \underline{82.06} & \underline{87.96} \\
 & LLaVA & FitNet~\cite{romero2014fitnets}    &    62.95   &    78.87   &  76.47     &   79.95   &  76.91    &  85.90    \\
 & LLaVA & CRD~\cite{tian2019contrastive}      &    \underline{69.47}  &    \underline{80.83}   &   \underline{80.93}    &   \underline{80.81}    &   81.80    &    86.86   \\
 & \cellcolor{gray!10}Qwen & \cellcolor{gray!10}\LaViD~(Ours)    & \cellcolor{gray!10}\textbf{70.15} & \cellcolor{gray!10}\textbf{81.51} & \cellcolor{gray!10}\textbf{81.34} & \cellcolor{gray!10}\textbf{83.22} & \cellcolor{gray!10}\textbf{82.29} & \cellcolor{gray!10}\textbf{88.59} \\

\midrule
\multirow{11}{*}{MNV2} & -- & Ind Student &   69.69    &  81.92  &  83.31   &   85.27    &  80.45  &  86.93 \\
 & RN-50 & KD~\cite{hinton2015distilling}      &   69.89 &   81.65   &    83.38   &   85.48    &   81.69    &    87.33   \\
 & RN-50 & RKD~\cite{park2019relational}      &  67.87     &   80.82    &   79.60    &    82.98   &    79.84   &  86.50     \\
 & RN-50 & DKD~\cite{zhao2022decoupled}      &    69.92   &   81.06  &   78.85    &   85.39    &   82.26   &    89.30   \\
 & RN-50 & MLKD~\cite{jin2023multi}     &    71.30   &    81.64   &   79.81    &   86.38    &   83.38    &   90.30    \\
  & RN-50 & LS~\cite{sun2024logit}     &    \underline{73.03}   &   \underline{84.38}    &    \underline{85.71}  &  \underline{87.96}   &    \underline{84.23}   &    \underline{91.27}  \\

   & \cellcolor{gray!10}QN+R50 & \cellcolor{gray!10}w/ \LaViD~(Ours)  &\cellcolor{gray!10}\textbf{75.62} & \cellcolor{gray!10}\textbf{85.01} & \cellcolor{gray!10}\textbf{87.74} &\cellcolor{gray!10}\textbf{88.08}   & \cellcolor{gray!10}\textbf{85.79} & \cellcolor{gray!10}\textbf{91.66}\\
   \cmidrule(lr){2-9}
 & InternVL & MaKD~\cite{lee2025multi}      & \underline{72.10}   &  \underline{82.46} &  \textbf{86.11}    &   83.81    &   83.48  &  \underline{86.83} \\
 & LLaVA & FitNet~\cite{romero2014fitnets}      &    70.00   &  82.27     &   83.46   &  \underline{85.46}    &    80.57   &     86.77 \\
 & LLaVA & CRD~\cite{tian2019contrastive}      &    71.96   &   80.28   &  85.21    &  82.31    &   \underline{83.87} &     86.59  \\
& \cellcolor{gray!10}Qwen & \cellcolor{gray!10}\LaViD~(Ours)  & \cellcolor{gray!10}\textbf{72.52}  & \cellcolor{gray!10}\textbf{84.22} & \cellcolor{gray!10}\underline{86.05}     &  \cellcolor{gray!10}\textbf{86.21}     &  \cellcolor{gray!10}\textbf{84.64}  & \cellcolor{gray!10}\textbf{87.93} \\
\midrule
\multirow{11}{*}{SNV2} & -- & Ind Student &   65.26   &  78.94  &  81.15   &   80.65   &   77.80  &  85.67 \\
 & RN-50 & KD~\cite{hinton2015distilling}      &    65.59  &   79.54   &   80.45    &    80.85  &   78.74    &    85.30   \\
 & RN-50 & RKD~\cite{park2019relational}      &    60.99   &   78.30    &   78.23    &  74.56    &   77.19    &     85.00 \\
 & RN-50 & DKD~\cite{zhao2022decoupled}      &    68.11  &    80.63   &   77.79    &   82.63   &   79.61    &   88.40    \\
 & RN-50 & MLKD~\cite{jin2023multi}     &    69.30   &    81.18  &   78.99    &   \textbf{84.12}    &    81.78   &   89.44   \\
  & RN-50 & LS~\cite{sun2024logit}     &   \underline{70.27}   &    \textbf{82.57}   &  \underline{83.71}     &    \underline{84.09}    &   \underline{83.23}    &   \underline{89.68}   \\
   & \cellcolor{gray!10}QN+R50 & \cellcolor{gray!10}w/ \LaViD~(Ours)   & \cellcolor{gray!10}\textbf{71.78}  & \cellcolor{gray!10}\underline{81.31} &  \cellcolor{gray!10}\textbf{85.23}  & \cellcolor{gray!10}83.72  &  \cellcolor{gray!10}\textbf{83.85}  &  \cellcolor{gray!10}\textbf{89.74} \\
     \cmidrule(lr){2-9}
 & InternVL & MaKD~\cite{lee2025multi}     & \underline{68.33}  &  \underline{79.84} &  \underline{82.69}    &   79.00    &   81.01  &  \underline{85.47} \\
 & LLaVA & FitNet~\cite{romero2014fitnets}      &   64.82    &  79.39     &   81.22   &  \underline{80.61}    &    77.78   &    85.04   \\
 & LLaVA & CRD~\cite{tian2019contrastive}      &   68.17    &   77.95   &   82.44    &    78.78  &     \textbf{81.28}  &    84.83   \\
 & \cellcolor{gray!10}Qwen &  \cellcolor{gray!10}\LaViD~(Ours)      & \cellcolor{gray!10}\textbf{68.53}  & \cellcolor{gray!10}\textbf{80.78} & \cellcolor{gray!10}\textbf{83.03}    & \cellcolor{gray!10}\textbf{81.63}   & \cellcolor{gray!10}\underline{81.19}  & \cellcolor{gray!10}\textbf{86.37} \\

\bottomrule
\end{tabular}
}
\caption{Top-1 (\%) accuracy of competing distillation approaches across six fine-grained classification benchmarks. RN-18, MNV2, and SNV2 denote ResNet-18, MobileNetV2, and ShuffleNetV2 student models, respectively. RN-50~\cite{he2016deep} serves as the conventional visual teacher, while InternVL~\cite{chen2024internvl} (InternVL2-8B), LLaVA~\cite{liu2023visual} (LLaVA-1.5-7B), and Qwen~\cite{yang2024qwen2} (Qwen2.5-7B) act as multimodal or language-only teachers. QN+R50 denotes a hybrid teacher setup combining Qwen2.5-7B and ResNet-50. The best and second-best results are marked in \textbf{bold} and \underline{underline}, respectively.}
\label{tab:distill-main-results}
\end{table*}

\begin{figure*}
    \centering
    \includegraphics[width=0.8\linewidth]{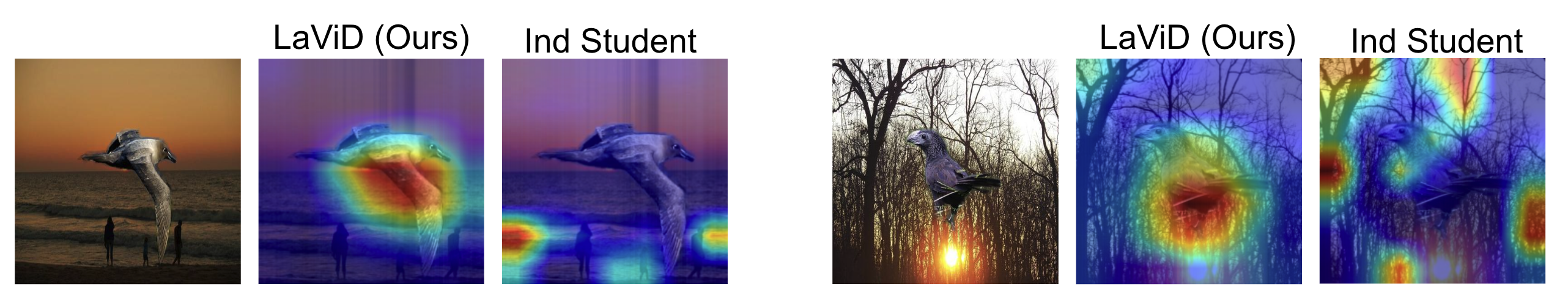}
    \caption{Grad-CAM visualizations on the Waterbirds dataset. \LaViD{} student better focuses on the bird rather than background artifacts.}
    \label{fig:waterbird-gradcam}
\end{figure*}

\subsection{Datasets and Implementation Details}
\textbf{Datasets.} We evaluate \LaViD{} on six fine-grained classification benchmarks: CUB-200~\cite{wah_branson_welinder_perona_belongie_2011}, Caltech-101~\cite{li_andreeto_ranzato_perona_2022}, 102Flowers~\cite{Nilsback08}, FGVC Aircraft~\cite{maji2013fine}, Oxford-IIIT Pet~\cite{parkhi2012cats}, and Stanford Cars~\cite{krause20133d}. These datasets naturally emphasize subtle visual distinctions between classes, making them well-suited for concept-driven supervision like ours. We further test \LaViD{} on large-scale datasets by evaluating it on ImageNet~\cite{deng2009imagenet}. Due to the limitations of our method in spanning large general classes, we do not run the full 1000-way classification task, but distinctively group into 9 semantically coherent subsets (e.g. birds, instruments) to assess the method's scalability. The full details are provided in the limitations and appendix.

\begin{table*}[h!]
\centering
\small
\begin{tabular}{lccccccc} 
\toprule
\textbf{Student} & \textbf{Method} & \textbf{CUB} & \textbf{Caltech} & \textbf{Flowers} & \textbf{Aircraft} & \textbf{Pets} & \textbf{Cars} \\ 
\midrule
\multirow{2}{*}{ViT} & Ind Student & 82.60 & 94.85 & 97.02 & 71.33 & 93.13 & 89.73 \\
  & \cellcolor{gray!10}\LaViD~(Ours) & \cellcolor{gray!10}\textbf{82.90} &  \cellcolor{gray!10}\textbf{95.05} &
  \cellcolor{gray!10}\textbf{97.28} &
  \cellcolor{gray!10}\textbf{78.08} & 
  \cellcolor{gray!10}\textbf{93.43} & 
  \cellcolor{gray!10}\textbf{89.73} \\
\midrule
\multirow{2}{*}{CLIP} & Ind Student & 84.65 & 92.13 & \textbf{98.32} & 81.41 & 94.45 & 92.33 \\
  & \cellcolor{gray!10}\LaViD~(Ours) & \cellcolor{gray!10}\textbf{86.69} & \cellcolor{gray!10}\textbf{92.51} & \cellcolor{gray!10}98.30 &
  \cellcolor{gray!10}\textbf{83.42} & 
  \cellcolor{gray!10}\textbf{94.49} &
  \cellcolor{gray!10}\textbf{93.06} \\
\bottomrule
\end{tabular}
\caption{Top-1 accuracy (\%) for ImageNet- and CLIP-pretrained ViT/B-16 models.}
\label{tab:vit-clip-students}
\end{table*}

\textbf{Implementation Details} We selected three well-studied student models for our main results: ResNet-18~\cite{he2016deep}, MobileNetV2~\cite{sandler2018mobilenetv2}, and ShuffleNetV2~\cite{ma2018shufflenet}. Unless otherwise specified, all experiments use Qwen2.5-7B~\cite{yang2024qwen2} as the language teacher in \LaViD{}, with MCQs generated by GPT-4o~\cite{openai2024gpt4o}. The full hyperparameters and training configurations are detailed in Appendix~\ref{subsec:trainingdetails}. All reported results are averaged over three trials.

\textbf{Baselines} We position our work within the broader context of knowledge distillation and compare \LaViD\ against several representative baselines. We include the following traditional distillation methods: KD~\cite{hinton2015distilling}, RKD~\cite{park2019relational}, DKD~\cite{zhao2022decoupled}, MLKD~\cite{jin2023multi}, and Logit Standardization (LS)~\cite{sun2024logit}. All these baselines require a dataset-specific teacher model trained on the same data as the student, making them effective in-domain. In addition, we compare with MaKD~\cite{lee2025multi}, a recent method that distills from multimodal large language models (MLLMs) by prompting with individual images. To further examine the effectiveness of MLLM-based supervision, we also adapt two feature-based distillation methods—CRD~\cite{tian2019contrastive} and FitNet~\cite{romero2014fitnets}—using LLaVA-1.5~\cite{liu2023visual} as the teacher. This establishes a more comprehensive multimodal feature-based baseline where student models are guided by MLLM-derived representations. Since MLLMs operate over token sequences, we extract features from multiple layers and find that middle layers (e.g., layer -12) tend to provide stronger supervision; full ablation results are provided in the Appendix.

\begin{table*}[t]
\centering
\small
\begin{tabular}{lcccccccc}
\toprule
\textbf{Method} & \textbf{Artifact} & \textbf{Bird} & \textbf{CONT} & \textbf{Device} & \textbf{INST} & \textbf{INV} & \textbf{Mammal} & \textbf{VERT} \\
\midrule
Ind Student & 73.40 & 88.65 & 69.54 & 70.98 & 72.98 & 76.61 & 78.55 & 69.97 \\
\cellcolor{gray!10}\LaViD\ (Ours) & \cellcolor{gray!10}\textbf{74.52} & \cellcolor{gray!10}\textbf{90.08} & \cellcolor{gray!10}\textbf{71.52} & \cellcolor{gray!10}\textbf{72.13} & \cellcolor{gray!10}\textbf{73.67} & \cellcolor{gray!10}\textbf{78.60} & \cellcolor{gray!10}\textbf{79.15} & \cellcolor{gray!10}\textbf{70.70} \\
\bottomrule
\end{tabular}
\caption{Top-1 accuracy (\%) with student ResNet-18 on ImageNet WordNet hierarchy synsets. CONT, INST, INV, VERT denote Container, Instrumentality, Invertebrate, and Vertebrate, respectively.}
\label{tab:imagenet}
\end{table*}

\subsection{Main Results}

Table~\ref{tab:distill-main-results} compares \LaViD{} with both traditional KD methods and recent approaches leveraging MLLMs as teachers. Notably, \LaViD{} consistently outperforms MLLM-based baselines, including MaKD~\cite{lee2025multi} and adaptations of FitNet~\cite{romero2014fitnets} and CRD~\cite{tian2019contrastive} with LLaVA~\cite{liu2023visual} as the teacher---despite our own language teacher (Qwen2.5-7B) never accessing training images. This demonstrates the effectiveness of conceptual supervision even in the absence of aligned multimodal data. 

\begin{figure*}
    \centering
    \includegraphics[width=0.8\linewidth]{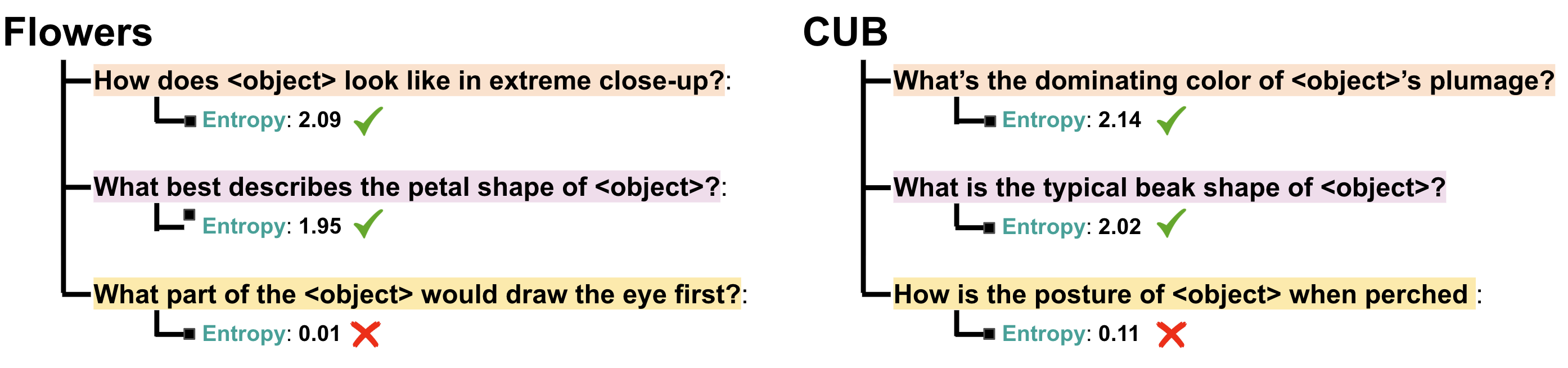}
\caption{Qualitative examples of high- and low-entropy questions on the Flowers and CUB datasets.}
\vspace{-0.5em}
\label{fig:question-entropy}
\end{figure*}

Our method also achieves competitive or superior performance compared to traditional visual teacher KD methods. For example, \LaViD{} surpasses DKD~\cite{zhao2022decoupled} and MLKD~\cite{jin2023multi} across several datasets such as CUB, Aircraft, Pets, and more. Furthermore, we find that our approach can be effectively combined with logit standardization (LS)~\cite{sun2024logit}, leading to additional gains across multiple datasets and student architectures.

We further evaluate \LaViD{} on subsets of ImageNet constructed using WordNet hierarchy synsets. As shown in Table~\ref{tab:imagenet}, \LaViD{} consistently outperforms the baseline across all 9 semantic groups. These results reinforce the effectiveness of language-derived supervision even at larger scale, without requiring access to multimodal data.

These results highlight that distillation from language-only teachers not only provides strong standalone performance, but also complements existing visual KD techniques—establishing \LaViD{} as a simple, modular, and broadly effective distillation paradigm.

To assess \LaViD{} beyond convolutional backbones, we evaluate its effectiveness on vision transformers, including ViT~\cite{dosovitskiy2020image} and CLIP~\cite{radford2021learning}, as student models in two variations: standard transformers initialized with ImageNet-pretrained weights~\cite{deng2009imagenet} to mitigate their data hunger, and CLIP models where we follow~\cite{wortsman2022robust} by initializing the classifier with the text embedding of “a photo of a {class}.” As shown in Table~\ref{tab:vit-clip-students}, \LaViD{} consistently improves over the baseline, demonstrating its applicability and generalizability to transformer-based models.

\begin{table*}[t]
\centering
\small
\begin{tabular}{lccccc} 
\toprule
\textbf{Student} & \textbf{Teacher} & \textbf{Method} & \textbf{Average} & \textbf{Best} & \textbf{Worst} \\ 
\midrule
\multirow{6}{*}{ResNet-18} & -- & Ind Student & \underline{72.81} & 98.36 & \underline{14.29} \\
                         & ResNet-50  & KD~\cite{hinton2015distilling} & 71.17 & 98.65 & 11.53 \\
                         & ResNet-50  & RKD~\cite{park2019relational} & 67.22 &  98.29  & 9.02 \\
                         & ResNet-50  & DKD~\cite{zhao2022decoupled} & 70.64 & 98.86 & 8.27 \\
                         & ResNet-50  & MLKD~\cite{jin2023multi} & 70.92 & \textbf{99.43} & 2.26 \\
                         & \cellcolor{gray!10}Qwen  & \cellcolor{gray!10}\LaViD~(Ours) & \cellcolor{gray!10}\textbf{86.10} & \cellcolor{gray!10}\underline{99.29}  & \cellcolor{gray!10}\textbf{55.39} \\
\midrule
\multirow{6}{*}{MobileNetV2} & -- & Ind Student & \underline{71.36} & 98.29  & \underline{18.05} \\
                         & ResNet-50  & KD~\cite{hinton2015distilling} & 71.00 & 98.57 & 15.54 \\
                         & ResNet-50  & RKD~\cite{park2019relational} & 67.53 & 98.43 & 8.02 \\
                         & ResNet-50  & DKD~\cite{zhao2022decoupled} & 71.00 & \underline{99.07} & 13.03 \\
                         & ResNet-50  & MLKD~\cite{jin2023multi} & 67.22 & \underline{99.07} & 3.51 \\
                         & \cellcolor{gray!10}Qwen  & \cellcolor{gray!10}\LaViD~(Ours) & \cellcolor{gray!10}\textbf{86.49} & \cellcolor{gray!10}\textbf{99.14} & \cellcolor{gray!10}\textbf{61.65} \\
\midrule
\multirow{6}{*}{ShuffleNetV2} & -- & Ind Student & \underline{72.28} & 98.07  & \underline{23.56} \\
                         & ResNet-50  & KD~\cite{hinton2015distilling} & 71.56 & 98.57 & 11.03 \\
                         & ResNet-50  & RKD~\cite{park2019relational} & 70.78 & \textbf{99.00} & 5.51 \\
                         & ResNet-50  & DKD~\cite{zhao2022decoupled} & 70.48 & 98.64 & 8.77 \\
                         & ResNet-50  & MLKD~\cite{jin2023multi} & 66.82 & \textbf{99.00} & 3.76 \\
                         & \cellcolor{gray!10}Qwen  & \cellcolor{gray!10}\LaViD~(Ours) & \cellcolor{gray!10}\textbf{84.26} & \cellcolor{gray!10}\underline{98.86} & \cellcolor{gray!10}\textbf{54.39} \\
\bottomrule
\end{tabular}
\caption{Top-1 accuracy (\%) of different distillation approaches evaluated on the Waterbirds dataset, grouped from the combination of \{waterbird, landbird\} and \{water background, land background\}. The best and second-best results are marked in \textbf{bold} and \underline{underline}, respectively. Average, Best, Worst represent the accuracy for each group.}
\vspace{-1em}
\label{tab:waterbird}
\end{table*}

\subsection{Overcoming Dataset Biases}
Prior work shows that student models can inherit biases from their teachers during knowledge distillation~\cite{ojha2023knowledge}. In contrast, \LaViD{} leverages general-purpose language models that are not trained on the visual data, providing supervision grounded in broad conceptual knowledge rather than dataset-specific patterns. This offers a unique opportunity to regularize the student with semantic guidance instead of spurious heuristics.

We validate this on Waterbirds~\cite{sagawa2019distributionally}, where spurious correlations between species and background make worst-group performance particularly challenging. As shown in Table~\ref{tab:waterbird}, \LaViD{} consistently achieves higher worst-group accuracy across all student architectures compared to independently trained models. The worst-performing group reflects biased models’ tendency to rely on background cues rather than relevant features, and traditional KD methods often exacerbate this issue by reinforcing shortcuts. \LaViD{}, however, mitigates these effects without compromising overall performance.

We provide further analysis with Grad-CAM in Figure~\ref{fig:waterbird-gradcam}, demonstrating \LaViD{} enforces student models to focus on the bird rather than spurious background elements.

\subsection{Analysis}
Unless otherwise specified, we conduct ablation studies using ResNet-18 on the CUB dataset. This configuration balances representative analysis with computational efficiency to analyze the core design choices in \LaViD{}.

\begin{figure}[ht]
\small
\centering
\includegraphics[scale=0.25]{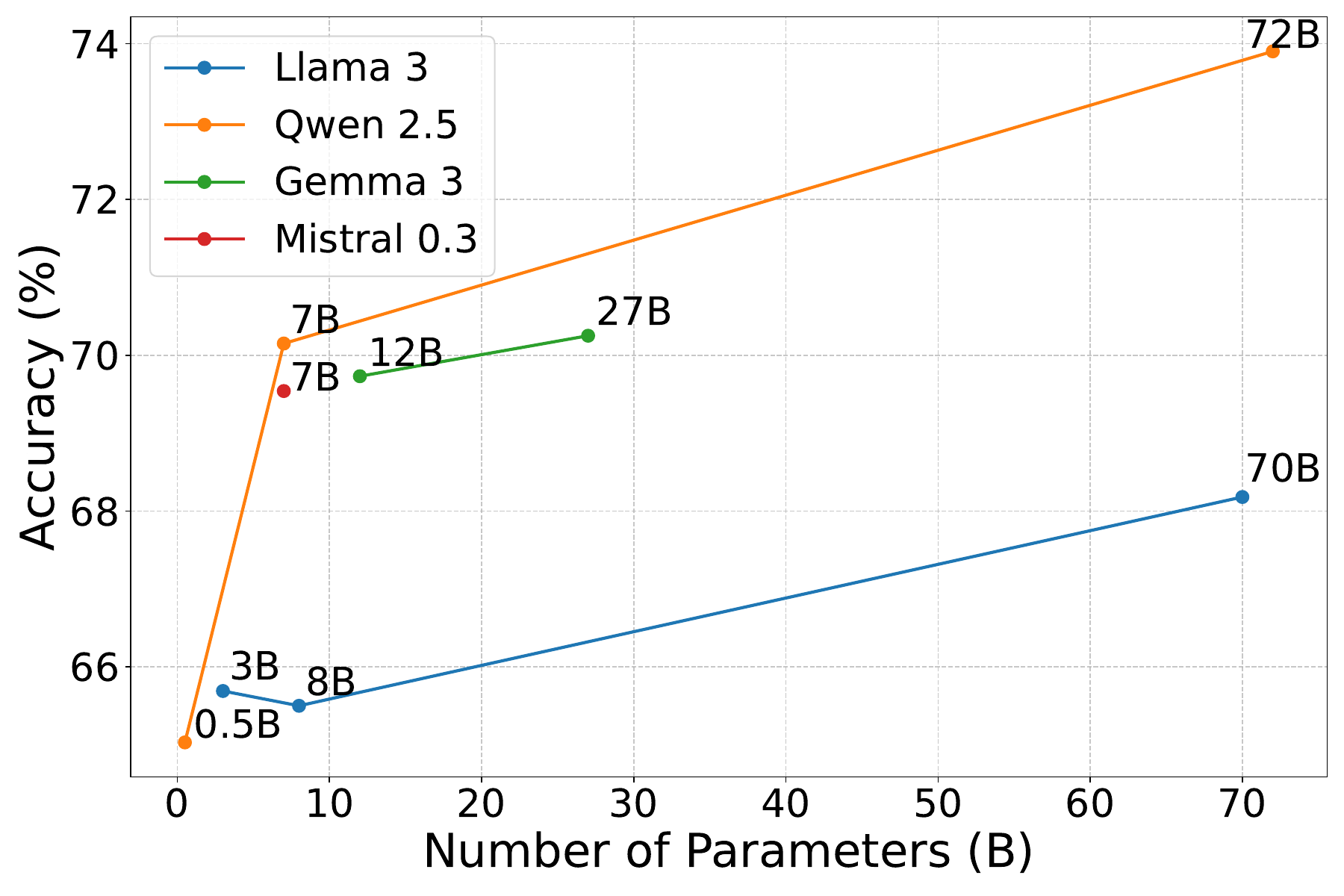}
\caption{Ablation on different LLM teachers with a ResNet-18 student on the CUB dataset.} 
\label{fig:llm_comparison_cub200}
\begin{tabular}{lcc} 
\toprule
\textbf{Student} & \textbf{Teacher} & \textbf{CUB}  \\ 
\midrule
\multirow{3}{*}{RN-18} 
  & MiniLM~\cite{wang2020minilm}  & 64.46  \\
  & BERT~\cite{devlin2019bert}    & 67.69 \\
  & \cellcolor{gray!10}\LaViD~(Ours)    & \cellcolor{gray!10}\textbf{70.15} \\
\midrule
\multirow{3}{*}{MNV2} 
  & MiniLM~\cite{wang2020minilm}  & 69.51  \\ 
  & BERT~\cite{devlin2019bert}    & 68.95  \\
  & \cellcolor{gray!10}\LaViD~(Ours)    & \cellcolor{gray!10}\textbf{72.52} \\
\midrule
\multirow{3}{*}{SNV2} 
  & MiniLM~\cite{wang2020minilm}  & 65.91  \\ 
  & BERT~\cite{devlin2019bert}    & 64.89  \\
  & \cellcolor{gray!10}\LaViD~(Ours)    & \cellcolor{gray!10}\textbf{68.53} \\
\bottomrule
\end{tabular}
\captionof{table}{Comparison between \LaViD{} and variants using static word embeddings (MiniLM, BERT) on the CUB dataset.}
\label{tab:ablation-word-embedding-mcq}
\end{figure}

\textbf{LLM vs. Word Embedding}
Since \LaViD{}’s MCQ supervision produces a logit vector that encodes inter-class relationships, we compare it with a word-embedding baseline that captures similar structure. In this variant, we directly use the pretrained word embedding of each class name from MiniLM~\cite{wang2020minilm} or BERT~\cite{devlin2019bert} as the supervision signal. As shown in Table~\ref{tab:ablation-word-embedding-mcq}, \LaViD{} consistently outperforms these word-embedding baselines, demonstrating that LLM-derived MCQs provide richer and more informative supervision than static embeddings.

\textbf{Choice of LLM Teachers}
In Figure~\ref{fig:llm_comparison_cub200}, we compare various LLM teachers within the \LaViD{} framework, including Qwen2.5 (0.5B, 7B, 70B), Gemma-3 (12B, 27B)~\cite{team2025gemma}, Mistral 0.3-7B~\cite{jiang2023mistral7b}, and LLaMA-3 (3B, 8B, 70B)~\cite{grattafiori2024llama}. Performance generally improves with model size, reflecting stronger semantic understanding. Qwen and Gemma consistently outperform LLaMA, and upon inspecting the logits, we find LLaMA’s are notably softer, which may limit its effectiveness as a teacher. We adopt Qwen2.5-7B for experiments as a tradeoff between performance and efficiency.

\textbf{Choice of MCQ generators.}
We further examine whether LaViD is sensitive to the LLM used for MCQ generation. Specifically, we replace GPT-4o~\cite{openai2024gpt4o} with Gemini 2.5 Pro~\cite{comanici2025gemini} for generating MCQs, while keeping Qwen-7B fixed as the LLM teacher for extracting logits. As shown in Table~\ref{tab:mcq_generator_ablation}, performance remains similar across the two MCQ generators, with only a small change on CUB and nearly identical performance on Caltech. This suggests that LaViD is not highly sensitive to the specific frontier LLM used to generate MCQs in our setting. This stability is consistent with our use of a constrained MCQ generation protocol, where the LLM is given the target class set and asked to generate questions that distinguish between these classes, rather than to freely propose labels or concepts.
\begin{table}[t]
\centering

\begin{tabular}{l l c c}
\toprule
\small
\textbf{MCQ Generator} & \textbf{LLM Teacher} & \textbf{CUB} & \textbf{Caltech} \\
\midrule
GPT-4o & Qwen-7B & 70.15 & 81.51 \\
Gemini 2.5 Pro & Qwen-7B & 69.25 & 81.64 \\
\bottomrule
\end{tabular}
\caption{
Effect of changing the LLM used for MCQ generation while keeping the LLM teacher consistent.
}
\label{tab:mcq_generator_ablation}
\end{table}

\begin{figure}[h]
    \centering 
    \begin{minipage}{0.4\textwidth} 
        \centering
        \includegraphics[width=\linewidth]{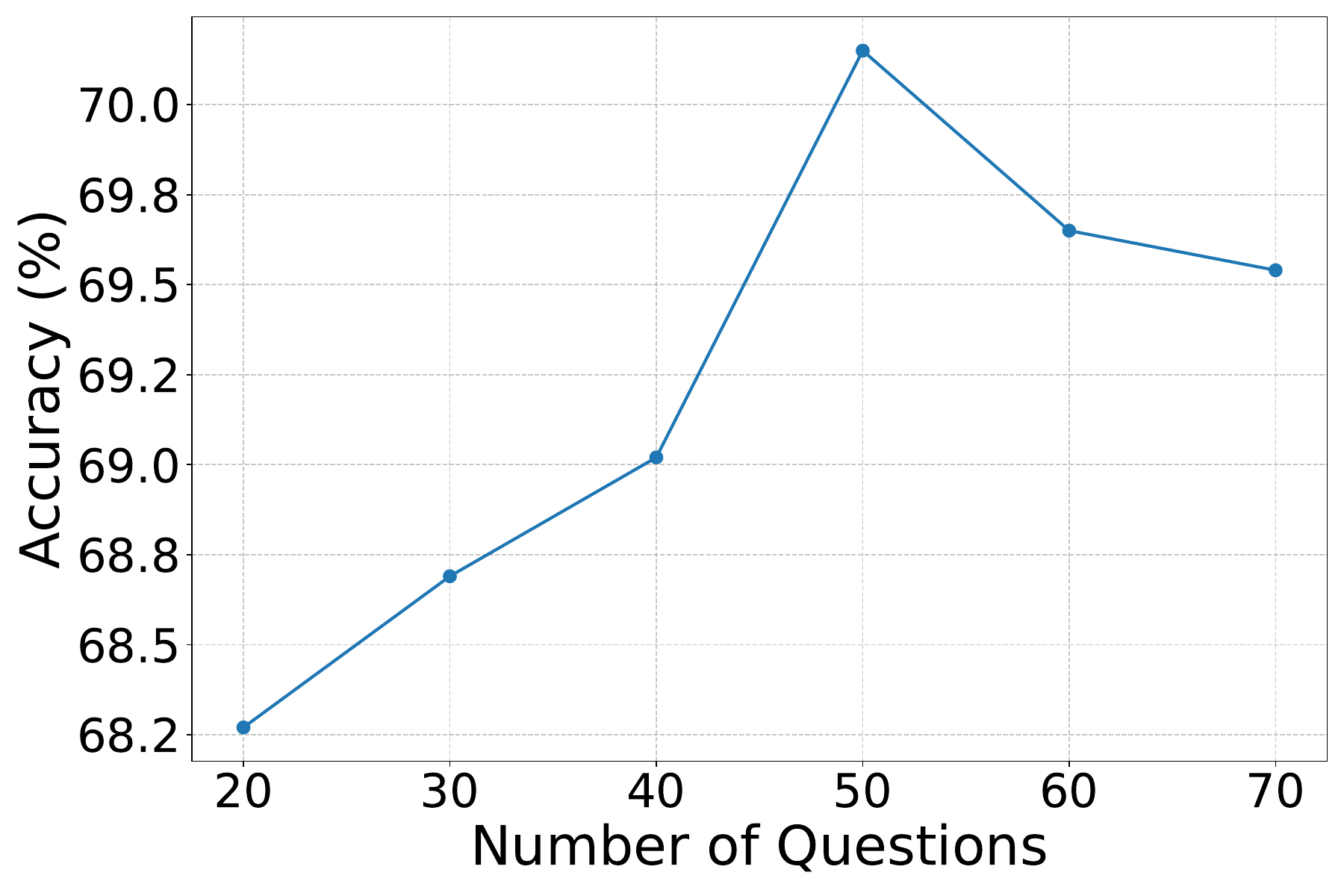}
        \caption{Effect of the number of MCQs (with 5 answer options) on accuracy of ResNet-18 student on CUB. Accuracy improves with more questions, but plateaus beyond 50.}
        \label{fig:ablation-question}
    \end{minipage}\hfill
    \begin{minipage}{0.4\textwidth}
        \centering
        \includegraphics[width=\linewidth]{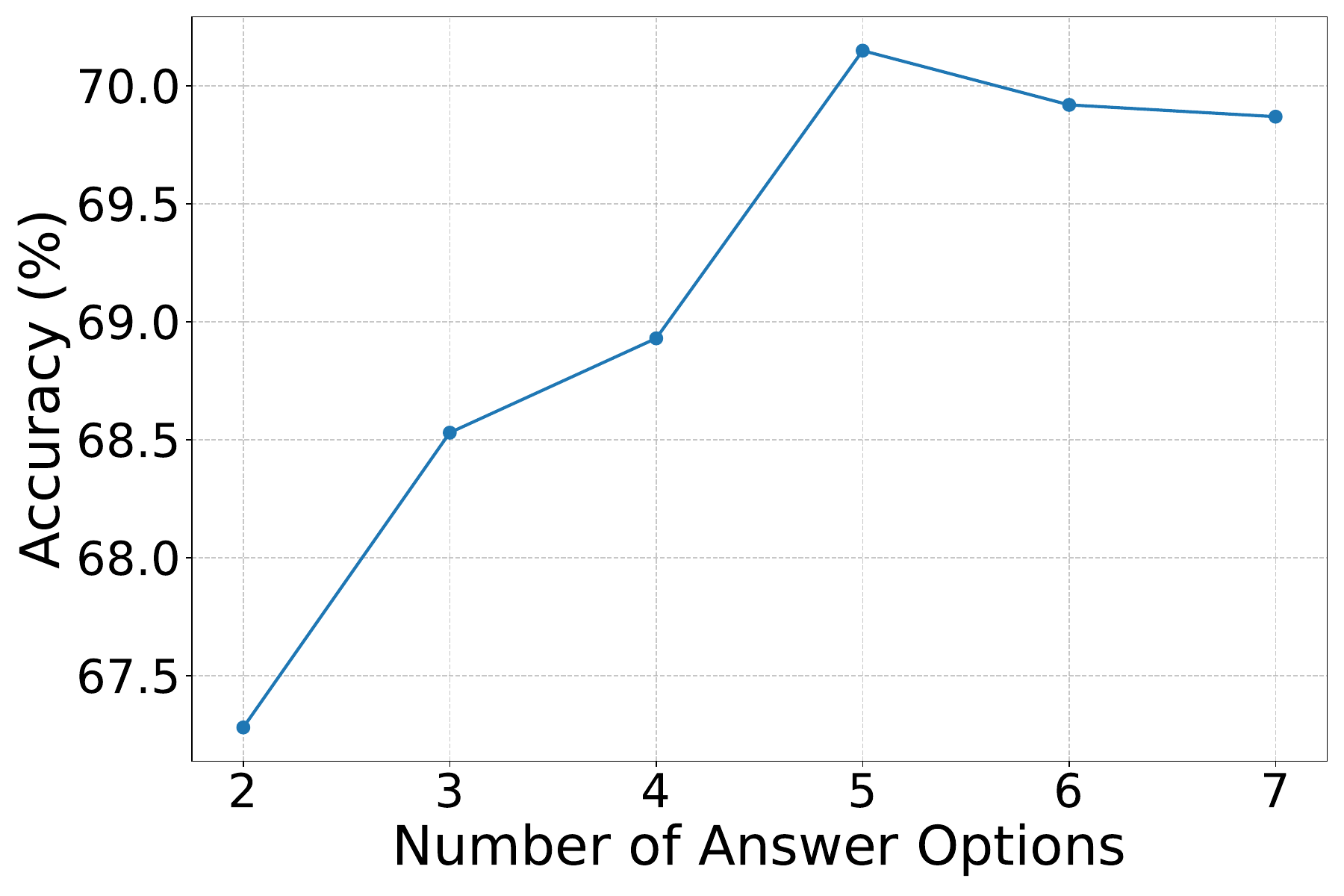}
        \caption{Effect of the number of answer options (with 50 questions) on CUB. Performance stabilizes after 5 options, highlighting question quality as a key factor.}
        \label{fig:ablation-answer}
    \end{minipage}
\end{figure}

\textbf{Number of Questions and Answer Options}
We investigate how the number of MCQs and their answer options affect distillation quality. Figure~\ref{fig:ablation-question} shows that increasing the number of questions improves student accuracy, but plateaus after 50 questions. This is likely because later questions degrade in semantic quality, as visually relevant distinctions become saturated. A similar trend is observed in Figure~\ref{fig:ablation-answer} for the number of answer options, where performance levels off beyond five. The effect is less pronounced, suggesting that question quality is more critical than granularity of choices. Based on these trends, we use 50 questions and 5 answer options in all experiments unless otherwise specified.

\textbf{Question Quality Analysis}
We measure discriminability via the average prediction entropy $\mathbb{E}_{c \in \mathcal{C}}[H(p_q^{(c)})]$. High entropy signals distinctive class semantics, while low entropy implies invariant attributes---both of which are shown in Fig.~\ref{fig:question-entropy}. These low-entropy questions are rare (fewer than 5\% of all questions), and removing them has negligible impact on accuracy, demonstrating that \LaViD{} remains robust even in the presence of less informative supervision.

\textbf{Additional Ablation Studies}
Further ablation studies including the effect of \LaViD\ loss weight and the number of questions and answer options, are provided in the Appendix.

Overall, although LaViD relies on language-model-generated questions to elicit conceptual structure, our ablations demonstrate that performance degrades gracefully under reduced question diversity and remains stable across different LLM backbones. This suggests that the method is not overly sensitive to individual prompt formulations but instead benefits from the aggregate semantic structure captured across multiple queries.

\section*{Limitations}

While \LaViD{} demonstrates strong performance across diverse fine-grained classification tasks, it has some limitations. The method relies on language-model-generated multiple-choice questions for conceptual supervision. The effectiveness of the supervision depends on the semantic coverage of the generated questions; however, our ablations show that performance degrades gracefully under reduced question diversity, suggesting that LaViD is not overly sensitive to individual prompt quality. In domains where distinctions are difficult to verbalize or LLMs lack domain familiarity, the conceptual signal may be less complete. Moreover, \LaViD{} assumes access to interpretable class names or metadata; in domains where class labels are abstract, underspecified, or not semantically meaningful, the approach may be less effective.

While our primary evaluation focuses on fine-grained recognition, which benefits most from external conceptual supervision, the proposed framework is agnostic to dataset size and model architecture. 
The main practical challenge in large-scale regimes lies in generating sufficiently diverse semantic queries to cover heterogeneous class spaces. 
Our subset experiments on ImageNet suggest that the conceptual supervision remains beneficial as class diversity increases, and scaling question generation strategies is a promising direction for future work. 
Notably, because supervision is generated once per dataset and reused throughout training, scaling to larger label spaces does not incur additional per-sample computational overhead.

\section{Conclusion}
In this work, we present \LaViD{}, a new paradigm for cross-modal knowledge distillation that transfers world knowledge from language-only large language models (LLMs) to vision-only student models. Our approach leverages multiple-choice questions generated from dataset metadata to extract structured semantic supervision without requiring paired image-text data or multimodal pretraining. Across six fine-grained benchmarks, we show that \LaViD{} consistently outperforms both traditional visual KD methods and recent multimodal approaches with a language-only teacher. Moreover, \LaViD{} shows strong robustness to spurious correlations and dataset biases, suggesting that external conceptual knowledge can steer student models toward more meaningful representations. Ablation studies further validate the importance of each component. Altogether, our findings establish \LaViD{} as a simple, effective, and general framework for infusing visual learners with high-level semantic understanding from language.

\section*{Impact Statement}

This work explores language-driven supervision for visual models, using language models to provide concept-level guidance through structured prompts. Rather than relying on language-vision pretraining or multimodal architectures, our method studies whether general-purpose knowledge encoded in LLMs can be distilled into visual learners as a complementary form of supervision.

Because LLMs are trained on broad and heterogeneous corpora, their outputs may reflect cultural assumptions, normative framing, or outdated knowledge. Even when supervision is provided through structured prompts rather than open-ended generation, these patterns may influence the resulting visual model. This work therefore highlights both the potential of large models as indirect teachers and the need to better understand the responsibilities and risks involved in transferring knowledge across modalities.

\section*{Acknowledgment}
This work was supported in part by NSF IIS2404180, and Institute of Information \& communications Technology Planning\& Evaluation (IITP) grants funded by the Korea government (MSIT) (No. 2022-0-00871, Development of AI Autonomy and Knowledge Enhancement for AI Agent Collaboration), and (No. RS-2025-2543949. Environment-Aware and Domain-Adaptive Multimodal Embodied AI for Real-World Interaction).

\bibliography{ref}

@inproceedings{buciluǎ2006model,
  title={Model compression},
  author={Buciluǎ, Cristian and Caruana, Rich and Niculescu-Mizil, Alexandru},
  booktitle={Proceedings of the 12th ACM SIGKDD international conference on Knowledge discovery and data mining},
  pages={535--541},
  year={2006}
}

@misc{schmidt2025saccadicvisionfinegrainedvisual,
      title={Saccadic Vision for Fine-Grained Visual Classification}, 
      author={Johann Schmidt and Sebastian Stober and Joachim Denzler and Paul Bodesheim},
      year={2025},
      eprint={2509.15688},
      archivePrefix={arXiv},
      primaryClass={cs.CV},
      url={https://arxiv.org/abs/2509.15688}, 
}

@misc{zheng2019learningdeepbilineartransformation,
      title={Learning Deep Bilinear Transformation for Fine-grained Image Representation}, 
      author={Heliang Zheng and Jianlong Fu and Zheng-Jun Zha and Jiebo Luo},
      year={2019},
      eprint={1911.03621},
      archivePrefix={arXiv},
      primaryClass={cs.CV},
      url={https://arxiv.org/abs/1911.03621}, 
}

@misc{shao2024datafreeknowledgedistillationfinegrained,
      title={Data-free Knowledge Distillation for Fine-grained Visual Categorization}, 
      author={Renrong Shao and Wei Zhang and Jianhua Yin and Jun Wang},
      year={2024},
      eprint={2404.12037},
      archivePrefix={arXiv},
      primaryClass={cs.CV},
      url={https://arxiv.org/abs/2404.12037}, 
}

@article{Wang_Wang_Li_Dou_Li_2020, title={Graph-Propagation Based Correlation Learning for Weakly Supervised Fine-Grained Image Classification}, volume={34}, url={https://ojs.aaai.org/index.php/AAAI/article/view/6912}, DOI={10.1609/aaai.v34i07.6912}, abstractNote={&lt;p&gt;The key of Weakly Supervised Fine-grained Image Classification (WFGIC) is how to pick out the discriminative regions and learn the discriminative features from them. However, most recent WFGIC methods pick out the discriminative regions independently and utilize their features directly, while neglecting the facts that regions’ features are mutually semantic correlated and region groups can be more discriminative. To address these issues, we propose an end-to-end Graph-propagation based Correlation Learning (GCL) model to fully mine and exploit the discriminative potentials of region correlations for WFGIC. Specifically, in discriminative region localization phase, a Criss-cross Graph Propagation (CGP) sub-network is proposed to learn region correlations, which establishes correlation between regions and then enhances each region by weighted aggregating other regions in a criss-cross way. By this means each region’s representation encodes the global image-level context and local spatial context simultaneously, thus the network is guided to implicitly discover the more powerful discriminative region groups for WFGIC. In discriminative feature representation phase, the Correlation Feature Strengthening (CFS) sub-network is proposed to explore the internal semantic correlation among discriminative patches’ feature vectors, to improve their discriminative power by iteratively enhancing informative elements while suppressing the useless ones. Extensive experiments demonstrate the effectiveness of proposed CGP and CFS sub-networks, and show that the GCL model achieves better performance both in accuracy and efficiency.&lt;/p&gt;}, number={07}, journal={Proceedings of the AAAI Conference on Artificial Intelligence}, author={Wang, Zhuhui and Wang, Shijie and Li, Haojie and Dou, Zhi and Li, Jianjun}, year={2020}, month={Apr.}, pages={12289-12296} }

@misc{ge2019weaklysupervisedcomplementaryparts,
      title={Weakly Supervised Complementary Parts Models for Fine-Grained Image Classification from the Bottom Up}, 
      author={Weifeng Ge and Xiangru Lin and Yizhou Yu},
      year={2019},
      eprint={1903.02827},
      archivePrefix={arXiv},
      primaryClass={cs.CV},
      url={https://arxiv.org/abs/1903.02827}, 
}

@misc{lin2017bilinearcnnsfinegrainedvisual,
      title={Bilinear CNNs for Fine-grained Visual Recognition}, 
      author={Tsung-Yu Lin and Aruni RoyChowdhury and Subhransu Maji},
      year={2017},
      eprint={1504.07889},
      archivePrefix={arXiv},
      primaryClass={cs.CV},
      url={https://arxiv.org/abs/1504.07889}, 
}

@article{hinton2015distilling,
  title={Distilling the knowledge in a neural network},
  author={Hinton, Geoffrey and Vinyals, Oriol and Dean, Jeff},
  journal={arXiv preprint arXiv:1503.02531},
  year={2015}
}

@article{huh2024platonic,
  title={The Platonic Representation Hypothesis},
  author={Minyoung Huh and Brian Cheung and Tongzhou Wang and Phillip Isola},
  journal={ICML},
  year={2024}
}

@article{romero2014fitnets,
  title={Fitnets: Hints for thin deep nets},
  author={Romero, Adriana and Ballas, Nicolas and Kahou, Samira Ebrahimi and Chassang, Antoine and Gatta, Carlo and Bengio, Yoshua},
  journal={arXiv preprint arXiv:1412.6550},
  year={2014}
}

@article{tian2019contrastive,
  title={Contrastive representation distillation},
  author={Tian, Yonglong and Krishnan, Dilip and Isola, Phillip},
  journal={arXiv preprint arXiv:1910.10699},
  year={2019}
}

@inproceedings{zhao2022decoupled,
  title={Decoupled knowledge distillation},
  author={Zhao, Borui and Cui, Quan and Song, Renjie and Qiu, Yiyu and Liang, Jiajun},
  booktitle={Proceedings of the IEEE/CVF Conference on computer vision and pattern recognition},
  pages={11953--11962},
  year={2022}
}

@inproceedings{chen2022knowledge,
  title={Knowledge distillation with the reused teacher classifier},
  author={Chen, Defang and Mei, Jian-Ping and Zhang, Hailin and Wang, Can and Feng, Yan and Chen, Chun},
  booktitle={Proceedings of the IEEE/CVF conference on computer vision and pattern recognition},
  pages={11933--11942},
  year={2022}
}

@inproceedings{yang2021knowledge,
  title={Knowledge distillation via softmax regression representation learning},
  author={Yang, Jing and Martinez, Brais and Bulat, Adrian and Tzimiropoulos, Georgios and others},
  year={2021},
  organization={International Conference on Learning Representations (ICLR)}
}

@article{hao2023one,
  title={One-for-all: Bridge the gap between heterogeneous architectures in knowledge distillation},
  author={Hao, Zhiwei and Guo, Jianyuan and Han, Kai and Tang, Yehui and Hu, Han and Wang, Yunhe and Xu, Chang},
  journal={Advances in Neural Information Processing Systems},
  volume={36},
  pages={79570--79582},
  year={2023}
}

@inproceedings{sun2024logit,
  title={Logit standardization in knowledge distillation},
  author={Sun, Shangquan and Ren, Wenqi and Li, Jingzhi and Wang, Rui and Cao, Xiaochun},
  booktitle={Proceedings of the IEEE/CVF conference on computer vision and pattern recognition},
  pages={15731--15740},
  year={2024}
}

@inproceedings{zhang2020improve,
  title={Improve object detection with feature-based knowledge distillation: Towards accurate and efficient detectors},
  author={Zhang, Linfeng and Ma, Kaisheng},
  booktitle={International conference on learning representations},
  year={2020}
}

@inproceedings{park2019relational,
  title={Relational knowledge distillation},
  author={Park, Wonpyo and Kim, Dongju and Lu, Yan and Cho, Minsu},
  booktitle={Proceedings of the IEEE/CVF conference on computer vision and pattern recognition},
  pages={3967--3976},
  year={2019}
}

@inproceedings{tung2019similarity,
  title={Similarity-preserving knowledge distillation},
  author={Tung, Frederick and Mori, Greg},
  booktitle={Proceedings of the IEEE/CVF international conference on computer vision},
  pages={1365--1374},
  year={2019}
}

@article{brown2020language,
  title={Language models are few-shot learners},
  author={Brown, Tom and Mann, Benjamin and Ryder, Nick and Subbiah, Melanie and Kaplan, Jared D and Dhariwal, Prafulla and Neelakantan, Arvind and Shyam, Pranav and Sastry, Girish and Askell, Amanda and others},
  journal={Advances in neural information processing systems},
  volume={33},
  pages={1877--1901},
  year={2020}
}

@article{raffel2020exploring,
  title={Exploring the limits of transfer learning with a unified text-to-text transformer},
  author={Raffel, Colin and Shazeer, Noam and Roberts, Adam and Lee, Katherine and Narang, Sharan and Matena, Michael and Zhou, Yanqi and Li, Wei and Liu, Peter J},
  journal={Journal of machine learning research},
  volume={21},
  number={140},
  pages={1--67},
  year={2020}
}

@article{chowdhery2023palm,
  title={Palm: Scaling language modeling with pathways},
  author={Chowdhery, Aakanksha and Narang, Sharan and Devlin, Jacob and Bosma, Maarten and Mishra, Gaurav and Roberts, Adam and Barham, Paul and Chung, Hyung Won and Sutton, Charles and Gehrmann, Sebastian and others},
  journal={Journal of Machine Learning Research},
  volume={24},
  number={240},
  pages={1--113},
  year={2023}
}

@article{touvron2023llama,
  title={Llama: Open and efficient foundation language models},
  author={Touvron, Hugo and Lavril, Thibaut and Izacard, Gautier and Martinet, Xavier and Lachaux, Marie-Anne and Lacroix, Timoth{\'e}e and Rozi{\`e}re, Baptiste and Goyal, Naman and Hambro, Eric and Azhar, Faisal and others},
  journal={arXiv preprint arXiv:2302.13971},
  year={2023}
}

@article{touvron2023llama2,
  title={Llama 2: Open foundation and fine-tuned chat models},
  author={Touvron, Hugo and Martin, Louis and Stone, Kevin and Albert, Peter and Almahairi, Amjad and Babaei, Yasmine and Bashlykov, Nikolay and Batra, Soumya and Bhargava, Prajjwal and Bhosale, Shruti and others},
  journal={arXiv preprint arXiv:2307.09288},
  year={2023}
}

@article{grattafiori2024llama,
  title={The llama 3 herd of models},
  author={Grattafiori, Aaron and Dubey, Abhimanyu and Jauhri, Abhinav and Pandey, Abhinav and Kadian, Abhishek and Al-Dahle, Ahmad and Letman, Aiesha and Mathur, Akhil and Schelten, Alan and Vaughan, Alex and others},
  journal={arXiv preprint arXiv:2407.21783},
  year={2024}
}

@article{yang2024qwen2,
  title={Qwen2. 5 technical report},
  author={Yang, An and Yang, Baosong and Zhang, Beichen and Hui, Binyuan and Zheng, Bo and Yu, Bowen and Li, Chengyuan and Liu, Dayiheng and Huang, Fei and Wei, Haoran and others},
  journal={arXiv preprint arXiv:2412.15115},
  year={2024}
}

@article{wang2020minilm,
  title={Minilm: Deep self-attention distillation for task-agnostic compression of pre-trained transformers},
  author={Wang, Wenhui and Wei, Furu and Dong, Li and Bao, Hangbo and Yang, Nan and Zhou, Ming},
  journal={Advances in neural information processing systems},
  volume={33},
  pages={5776--5788},
  year={2020}
}

@inproceedings{radford2021learning,
  title={Learning transferable visual models from natural language supervision},
  author={Radford, Alec and Kim, Jong Wook and Hallacy, Chris and Ramesh, Aditya and Goh, Gabriel and Agarwal, Sandhini and Sastry, Girish and Askell, Amanda and Mishkin, Pamela and Clark, Jack and others},
  booktitle={International conference on machine learning},
  pages={8748--8763},
  year={2021},
  organization={PmLR}
}

@inproceedings{devlin2019bert,
  title={Bert: Pre-training of deep bidirectional transformers for language understanding},
  author={Devlin, Jacob and Chang, Ming-Wei and Lee, Kenton and Toutanova, Kristina},
  booktitle={Proceedings of the 2019 conference of the North American chapter of the association for computational linguistics: human language technologies, volume 1 (long and short papers)},
  pages={4171--4186},
  year={2019}
}

@article{liu2023visual,
  title={Visual instruction tuning},
  author={Liu, Haotian and Li, Chunyuan and Wu, Qingyang and Lee, Yong Jae},
  journal={Advances in neural information processing systems},
  volume={36},
  pages={34892--34916},
  year={2023}
}

@article{ojha2023knowledge,
  title={What knowledge gets distilled in knowledge distillation?},
  author={Ojha, Utkarsh and Li, Yuheng and Sundara Rajan, Anirudh and Liang, Yingyu and Lee, Yong Jae},
  journal={Advances in Neural Information Processing Systems},
  volume={36},
  pages={11037--11048},
  year={2023}
}

@article{huang2022knowledge,
  title={Knowledge distillation from a stronger teacher},
  author={Huang, Tao and You, Shan and Wang, Fei and Qian, Chen and Xu, Chang},
  journal={Advances in Neural Information Processing Systems},
  volume={35},
  pages={33716--33727},
  year={2022}
}

@article{fan2024scalekd,
  title={ScaleKD: Strong Vision Transformers Could Be Excellent Teachers},
  author={Fan, Jiawei and Li, Chao and Liu, Xiaolong and Yao, Anbang},
  journal={arXiv preprint arXiv:2411.06786},
  year={2024}
}

@inproceedings{zhu2023good,
  title={A good student is cooperative and reliable: Cnn-transformer collaborative learning for semantic segmentation},
  author={Zhu, Jinjing and Luo, Yunhao and Zheng, Xu and Wang, Hao and Wang, Lin},
  booktitle={Proceedings of the IEEE/CVF International Conference on Computer Vision},
  pages={11720--11730},
  year={2023}
}

@inproceedings{liu2022cross,
  title={Cross-architecture knowledge distillation},
  author={Liu, Yufan and Cao, Jiajiong and Li, Bing and Hu, Weiming and Ding, Jingting and Li, Liang},
  booktitle={Proceedings of the Asian conference on computer vision},
  pages={3396--3411},
  year={2022}
}

@article{xue2022modality,
  title={The modality focusing hypothesis: Towards understanding crossmodal knowledge distillation},
  author={Xue, Zihui and Gao, Zhengqi and Ren, Sucheng and Zhao, Hang},
  journal={arXiv preprint arXiv:2206.06487},
  year={2022}
}

@article{gu2021open,
  title={Open-vocabulary object detection via vision and language knowledge distillation},
  author={Gu, Xiuye and Lin, Tsung-Yi and Kuo, Weicheng and Cui, Yin},
  journal={arXiv preprint arXiv:2104.13921},
  year={2021}
}

@inproceedings{wu2023aligning,
  title={Aligning bag of regions for open-vocabulary object detection},
  author={Wu, Size and Zhang, Wenwei and Jin, Sheng and Liu, Wentao and Loy, Chen Change},
  booktitle={Proceedings of the IEEE/CVF conference on computer vision and pattern recognition},
  pages={15254--15264},
  year={2023}
}

@inproceedings{xu2023masqclip,
  title={Masqclip for open-vocabulary universal image segmentation},
  author={Xu, Xin and Xiong, Tianyi and Ding, Zheng and Tu, Zhuowen},
  booktitle={Proceedings of the IEEE/CVF International Conference on Computer Vision},
  pages={887--898},
  year={2023}
}

@inproceedings{xue2021multimodal,
  title={Multimodal knowledge expansion},
  author={Xue, Zihui and Ren, Sucheng and Gao, Zhengqi and Zhao, Hang},
  booktitle={Proceedings of the IEEE/CVF International Conference on Computer Vision},
  pages={854--863},
  year={2021}
}

@inproceedings{gupta2016cross,
  title={Cross modal distillation for supervision transfer},
  author={Gupta, Saurabh and Hoffman, Judy and Malik, Jitendra},
  booktitle={Proceedings of the IEEE conference on computer vision and pattern recognition},
  pages={2827--2836},
  year={2016}
}

@inproceedings{garcia2018modality,
  title={Modality distillation with multiple stream networks for action recognition},
  author={Garcia, Nuno C and Morerio, Pietro and Murino, Vittorio},
  booktitle={Proceedings of the European Conference on Computer Vision (ECCV)},
  pages={103--118},
  year={2018}
}

@article{wah_branson_welinder_perona_belongie_2011, title={The Caltech-UCSD Birds-200-2011 Dataset}, abstractNote={CUB-200-2011 is an extended version of CUB-200 [7], a challenging dataset of 200 bird species. The extended version roughly doubles the number of images per category and adds new part localization annotations. All images are annotated with bounding boxes, part locations, and at- tribute labels. Images and annotations were filtered by mul- tiple users of Mechanical Turk. We introduce benchmarks and baseline experiments for multi-class categorization and part localization.}, publisher={California Institute of Technology}, author={Wah, Catherine and Branson, Steve and Welinder, Peter and Perona, Pietro and Belongie, Serge}, year={2011}, month={Jul} }

@misc{li_andreeto_ranzato_perona_2022, title={Caltech 101}, DOI={10.22002/D1.20086}, abstractNote={Pictures of objects belonging to 101 categories. About 40 to 800 images per category. Most categories have about 50 images. Collected in September 2003 by Fei-Fei Li, Marco Andreetto, and Marc'Aurelio Ranzato. The size of each image is roughly 300 x 200 pixels. We have carefully clicked outlines of each object in these pictures, these are included under the 'Annotations.tar'. There is also a MATLAB script to view the annotations, 'show_annotations.m'.}, publisher={CaltechDATA}, author={Li, Fei-Fei and Andreeto, Marco and Ranzato, Marc'Aurelio and Perona, Pietro}, year={2022}, month={Apr} }

@InProceedings{Nilsback08,
  author       = "Maria-Elena Nilsback and Andrew Zisserman",
  title        = "Automated Flower Classification over a Large Number of Classes",
  booktitle    = "Indian Conference on Computer Vision, Graphics and Image Processing",
  month        = "Dec",
  year         = "2008",
}

@article{maji2013fine,
  title={Fine-grained visual classification of aircraft},
  author={Maji, Subhransu and Rahtu, Esa and Kannala, Juho and Blaschko, Matthew and Vedaldi, Andrea},
  journal={arXiv preprint arXiv:1306.5151},
  year={2013}
}

@inproceedings{parkhi2012cats,
  title={Cats and dogs},
  author={Parkhi, Omkar M and Vedaldi, Andrea and Zisserman, Andrew and Jawahar, CV},
  booktitle={2012 IEEE conference on computer vision and pattern recognition},
  pages={3498--3505},
  year={2012},
  organization={IEEE}
}

@inproceedings{krause20133d,
  title={3d object representations for fine-grained categorization},
  author={Krause, Jonathan and Stark, Michael and Deng, Jia and Fei-Fei, Li},
  booktitle={Proceedings of the IEEE international conference on computer vision workshops},
  pages={554--561},
  year={2013}
}

@inproceedings{deng2009imagenet,
  title={Imagenet: A large-scale hierarchical image database},
  author={Deng, Jia and Dong, Wei and Socher, Richard and Li, Li-Jia and Li, Kai and Fei-Fei, Li},
  booktitle={2009 IEEE conference on computer vision and pattern recognition},
  pages={248--255},
  year={2009},
  organization={Ieee}
}

@article{sagawa2019distributionally,
  title={Distributionally robust neural networks for group shifts: On the importance of regularization for worst-case generalization},
  author={Sagawa, Shiori and Koh, Pang Wei and Hashimoto, Tatsunori B and Liang, Percy},
  journal={arXiv preprint arXiv:1911.08731},
  year={2019}
}

@article{lee2025multi,
  title={Multi-aspect Knowledge Distillation with Large Language Model},
  author={Lee, Taegyeong and Bang, Jinsik and Kwon, Soyeong and Kim, Taehwan},
  journal={arXiv preprint arXiv:2501.13341},
  year={2025}
}

@inproceedings{ma2018shufflenet,
  title={Shufflenet v2: Practical guidelines for efficient cnn architecture design},
  author={Ma, Ningning and Zhang, Xiangyu and Zheng, Hai-Tao and Sun, Jian},
  booktitle={Proceedings of the European conference on computer vision (ECCV)},
  pages={116--131},
  year={2018}
}

@inproceedings{sandler2018mobilenetv2,
  title={Mobilenetv2: Inverted residuals and linear bottlenecks},
  author={Sandler, Mark and Howard, Andrew and Zhu, Menglong and Zhmoginov, Andrey and Chen, Liang-Chieh},
  booktitle={Proceedings of the IEEE conference on computer vision and pattern recognition},
  pages={4510--4520},
  year={2018}
}

@inproceedings{he2016deep,
  title={Deep residual learning for image recognition},
  author={He, Kaiming and Zhang, Xiangyu and Ren, Shaoqing and Sun, Jian},
  booktitle={Proceedings of the IEEE conference on computer vision and pattern recognition},
  pages={770--778},
  year={2016}
}

@inproceedings{jin2023multi,
  title={Multi-level logit distillation},
  author={Jin, Ying and Wang, Jiaqi and Lin, Dahua},
  booktitle={Proceedings of the IEEE/CVF Conference on Computer Vision and Pattern Recognition},
  pages={24276--24285},
  year={2023}
}

@article{steiner2021train,
  title={How to train your vit? data, augmentation, and regularization in vision transformers},
  author={Steiner, Andreas and Kolesnikov, Alexander and Zhai, Xiaohua and Wightman, Ross and Uszkoreit, Jakob and Beyer, Lucas},
  journal={arXiv preprint arXiv:2106.10270},
  year={2021}
}

@article{team2025gemma,
  title={Gemma 3 technical report},
  author={Team, Gemma and Kamath, Aishwarya and Ferret, Johan and Pathak, Shreya and Vieillard, Nino and Merhej, Ramona and Perrin, Sarah and Matejovicova, Tatiana and Ram{\'e}, Alexandre and Rivi{\`e}re, Morgane and others},
  journal={arXiv preprint arXiv:2503.19786},
  year={2025}
}

@article{dosovitskiy2020image,
  title={An image is worth 16x16 words: Transformers for image recognition at scale},
  author={Dosovitskiy, Alexey and Beyer, Lucas and Kolesnikov, Alexander and Weissenborn, Dirk and Zhai, Xiaohua and Unterthiner, Thomas and Dehghani, Mostafa and Minderer, Matthias and Heigold, Georg and Gelly, Sylvain and others},
  journal={arXiv preprint arXiv:2010.11929},
  year={2020}
}

@inproceedings{chen2024internvl,
  title={Internvl: Scaling up vision foundation models and aligning for generic visual-linguistic tasks},
  author={Chen, Zhe and Wu, Jiannan and Wang, Wenhai and Su, Weijie and Chen, Guo and Xing, Sen and Zhong, Muyan and Zhang, Qinglong and Zhu, Xizhou and Lu, Lewei and others},
  booktitle={Proceedings of the IEEE/CVF conference on computer vision and pattern recognition},
  pages={24185--24198},
  year={2024}
}

@misc{openai2024gpt4o,
  title        = {GPT-4o System Card},
  author       = {{OpenAI}},
  year         = {2024},
  howpublished = {\url{https://arxiv.org/abs/2410.21276}},
  note         = {Accessed: 2025-05-16}
}

@inproceedings{wortsman2022robust,
  title={Robust fine-tuning of zero-shot models},
  author={Wortsman, Mitchell and Ilharco, Gabriel and Kim, Jong Wook and Li, Mike and Kornblith, Simon and Roelofs, Rebecca and Lopes, Raphael Gontijo and Hajishirzi, Hannaneh and Farhadi, Ali and Namkoong, Hongseok and others},
  booktitle={Proceedings of the IEEE/CVF conference on computer vision and pattern recognition},
  pages={7959--7971},
  year={2022}
}

@misc{jiang2023mistral7b,
      title={Mistral 7B}, 
      author={Albert Q. Jiang and Alexandre Sablayrolles and Arthur Mensch and Chris Bamford and Devendra Singh Chaplot and Diego de las Casas and Florian Bressand and Gianna Lengyel and Guillaume Lample and Lucile Saulnier and Lélio Renard Lavaud and Marie-Anne Lachaux and Pierre Stock and Teven Le Scao and Thibaut Lavril and Thomas Wang and Timothée Lacroix and William El Sayed},
      year={2023},
      eprint={2310.06825},
      archivePrefix={arXiv},
      primaryClass={cs.CL},
      url={https://arxiv.org/abs/2310.06825}, 
}

@article{comanici2025gemini,
  title={Gemini 2.5: Pushing the frontier with advanced reasoning, multimodality, long context, and next generation agentic capabilities},
  author={Comanici, Gheorghe and Bieber, Eric and Schaekermann, Mike and Pasupat, Ice and Sachdeva, Noveen and Dhillon, Inderjit and Blistein, Marcel and Ram, Ori and Zhang, Dan and Rosen, Evan and others},
  journal={arXiv preprint arXiv:2507.06261},
  year={2025}
}
\bibliographystyle{icml2026}

\newpage
\appendix
\onecolumn
\appendix
\section*{Appendix}
\label{sec:appendix}

\setcounter{figure}{0}
\setcounter{table}{0}
\setcounter{section}{0}

\renewcommand{\theequation}{\Alph{equation}}
\renewcommand{\thefigure}{\Alph{figure}}
\renewcommand{\thesection}{\Alph{section}}
\renewcommand{\thetable}{\Alph{table}}

\section{Additional Experimental Information}

\subsection{Dataset Details}
\label{sec:dataset-details}

We conduct experiments on a diverse set of widely used fine-grained visual classification benchmarks. Below, we provide a brief description of each dataset used in our evaluation.

\paragraph{Stanford Cars}~\cite{krause20133d}  
The Stanford Cars dataset comprises 16,185 images across 196 fine-grained car categories defined by make, model, and year (e.g., \textit{2012 Tesla Model S}, \textit{2012 BMW M3 Coupe}). The data are split into 8,144 training and 8,041 testing samples, with each class approximately balanced between the two splits.

\paragraph{Oxford Pets}~\cite{parkhi2012cats}  
Oxford Pets contains 7,384 images of 37 cat and dog breeds, with roughly 200 samples per class. The dataset is divided into 3,690 training and 3,694 testing images and is characterized by significant variability in scale, pose, lighting, and appearance, making it a useful benchmark for robust recognition.

\paragraph{102 Flowers}~\cite{Nilsback08}  
The 102 Flowers dataset includes images of 102 flower species, with 6,552 images used for training and 1,637 for testing. Each class contains between 40 and 258 samples, exhibiting considerable diversity in viewpoint, scale, and illumination conditions.

\paragraph{CUB-200}~\cite{wah_branson_welinder_perona_belongie_2011}  
CUB-200 is a standard benchmark for fine-grained visual categorization, consisting of 11,788 bird images spanning 200 species. The dataset is split into 5,994 training and 5,794 testing samples. Each image is annotated with rich metadata, including part locations, binary attributes, and bounding boxes, enabling more detailed analysis beyond classification accuracy.

\paragraph{FGVC-Aircraft}~\cite{maji2013fine}  
FGVC-Aircraft comprises 9,967 images covering 100 aircraft model variants, with around 100 samples per class. The dataset is divided into 6,667 training and 3,300 testing images. Each image is provided with a tight bounding box and a four-level hierarchical label describing the aircraft type and model.

\paragraph{Caltech-101}~\cite{li_andreeto_ranzato_perona_2022}  
Caltech-101 contains images from 101 object categories. To focus exclusively on object classification, the background category included in the original release is excluded. The dataset consists of 4,310 training and 4,367 testing images, with most categories containing around 50 samples.

\subsection{Training Details}
\label{subsec:trainingdetails}

\begin{figure}
\centering
\includegraphics[scale=0.25]{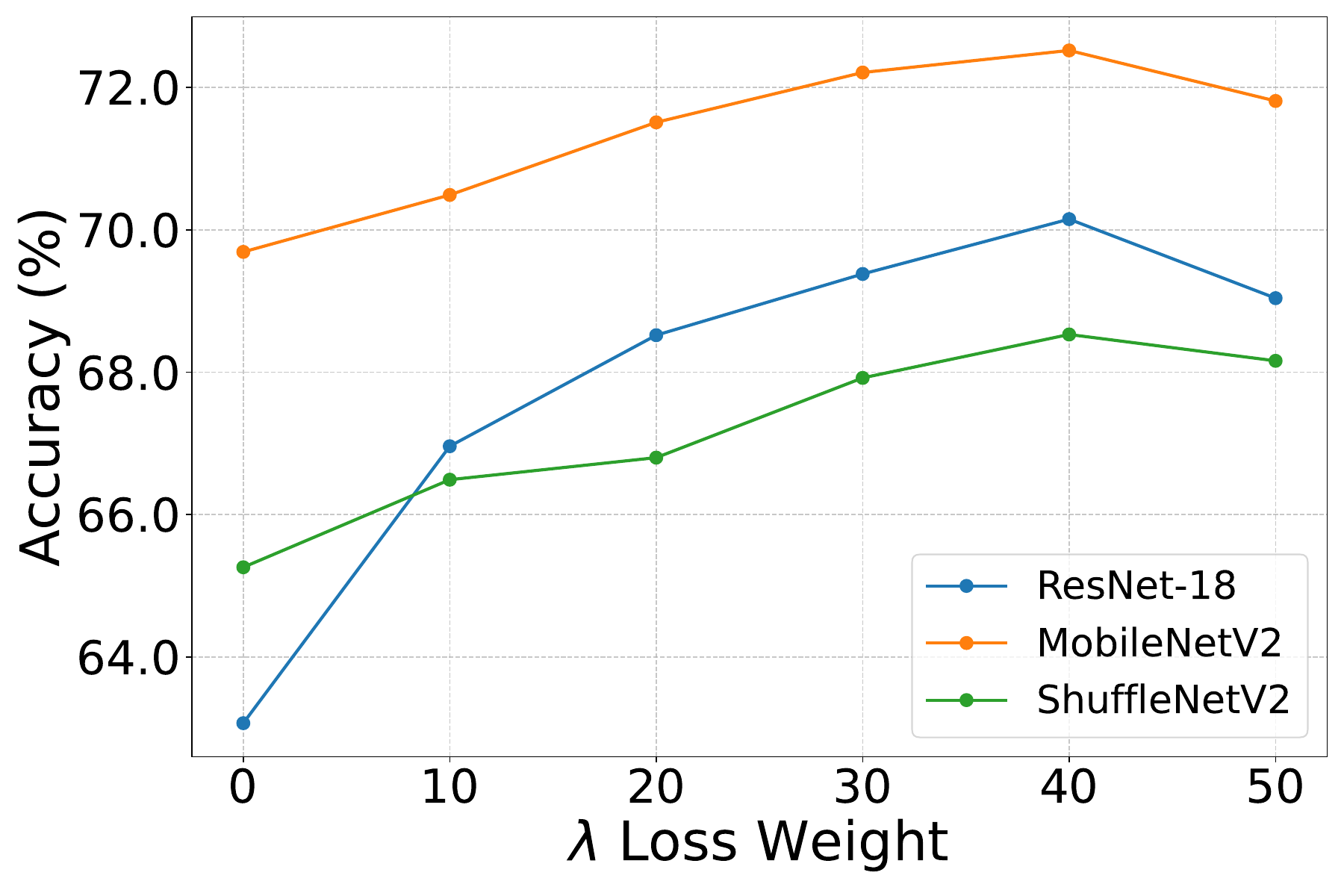}
\caption{Effect of loss weight \( \lambda \) on student accuracy for ResNet-18, MobileNetV2, and ShuffleNetV2 on the CUB dataset.}
\label{fig:loss_weights_cub200}
\end{figure}

On all datasets apart from ImageNet, we train the student models for 240 epochs with a batch size of 16. The initial learning rate is 0.01, divided by 10 at epochs 150, 180, and 210. The optimizer is SGD with a momentum of 0.9 and weight decay of 5e-4. For the traditional KD methods, the teachers are trained with the same hyperparameters. On ImageNet, we follow standard practices: train all models for 100 epochs with a batch size of 512 on 8 GPUs. The initial learning rate is 0.2, divided by 10 at epoch 30, 60, and 90. ViT models are fine-tuned for 500 epochs with a batch size of 512 using a cosine learning rate scheduler~\cite{steiner2021train}, while CLIP models are fine-tuned for 75 epochs with a batch size of 16. The LaViD loss weight $\lambda$ for experiments is presented in Table~\ref{tab:student_dataset_loss_weight}. We also show the impact of loss weight using CUB dataset as an example (Figure~\ref{fig:loss_weights_cub200}).

\begin{table*}[ht]
\small
\centering
\begin{tabular}{lcccccc}
\toprule
\textbf{Student} & \textbf{CUB} & \textbf{Caltech} & \textbf{Flowers} & \textbf{Aircraft} & \textbf{Pets} & \textbf{Cars} \\
\midrule
ResNet-18 & 40.0 & 20.0 & 50.0 & 50.0 & 50.0 & 40.0 \\
MobileNetV2 & 40.0 & 20.0 & 30.0 & 20.0 & 20.0 & 30.0 \\
ShuffleNetV2 & 40.0 & 20.0 & 30.0 & 10.0 & 30.0 & 20.0 \\
\bottomrule
\end{tabular}
\caption{LaViD loss weight used in training $\lambda$ (tested with intervals of 10).}
\label{tab:student_dataset_loss_weight}
\end{table*}

\subsection{Computational Resources}
\label{subsec:compute}

All experiments in this work were conducted using consumer-grade GPUs, specifically NVIDIA GeForce RTX~2080~Ti or RTX~3090, depending on resource availability at the time of training. Due to occasional hardware scheduling differences, the actual GPU hours may vary slightly across runs. Nevertheless, the total training time for each experiment remained relatively small — typically on the order of only a few hours — reflecting the lightweight nature of the models and tasks considered in this study.

\subsection{ImageNet Hierarchy Synsets}

As described in Section 4.2, we evaluate \LaViD{} on nine separate subsets of ImageNet, each corresponding to a semantic group defined by the WordNet hierarchy. These include domains such as \textit{mammal}, \textit{bird}, and \textit{artifact}. Out of the 1,000 ImageNet classes, 858 are associated with WordNet synsets that fall into one of these groups. Below, we provide the full list of classes for each group, corresponding to the results reported in Table~\ref{tab:imagenet}.

\underline{Artifact:} altar, apiary, bakery, Band-Aid, baluster / handrail, barbershop, barn, bath towel, lighthouse, bell tower, baby bib, ring binder, birdhouse, boathouse, bookstore, bottle cap, brass memorial plaque, breakwater, breastplate, butcher shop, castle, chain-link fence, chain mail, church, movie theater, cliff dwelling, cloak, clogs, spiral or coil, candy store, cowboy boot, cuirass, dam, dishcloth, dock, dome, doormat, fire screen, fountain, gas mask or respirator, greenhouse, radiator grille, grocery store, handkerchief, holster, home theater, honeycomb, lampshade, lens cap, library, slip-on shoe, sawmill, manhole cover, megalith, monastery, mosque, mosquito net, tent, necklace, baby pacifier, obelisk, palace, paper towel, patio, pedestal, Pickelhaube, picket fence, pillow, planetarium, plate rack, prison, quilt, restaurant, sneaker, sandal, scabbard, scoreboard, shield, shoe store, shoji screen / room divider, balaclava ski mask, sliding door, stage, through arch bridge, stone wall, stupa, suspension bridge, teddy bear, thatched roof, tile roof, tobacco shop, totem pole, toy store, triumphal arch, turnstile, umbrella, vaulted or arched ceiling, velvet fabric, viaduct, window screen, window shade, wool, split-rail fence, yurt, dust jacket

\underline{Bird:} rooster, hen, ostrich, brambling, goldfinch, house finch, junco, indigo bunting, American robin, bulbul, jay, magpie, chickadee, American dipper, kite (bird of prey), bald eagle, vulture, great grey owl, black grouse, ptarmigan, ruffed grouse, prairie grouse, peafowl, quail, partridge, african grey parrot, macaw, sulphur-crested cockatoo, lorikeet, coucal, bee eater, hornbill, hummingbird, jacamar, toucan, duck, red-breasted merganser, goose, black swan, white stork, black stork, spoonbill, flamingo, little blue heron, great egret, bittern bird, crane bird, limpkin, common gallinule, American coot, bustard, ruddy turnstone, dunlin, common redshank, dowitcher, oystercatcher, pelican, king penguin, albatross

\underline{Container:} ambulance, amphibious vehicle, trash can, backpack, barrel, wheelbarrow, bathtub, station wagon, beaker, beer bottle, beer glass, tandem bicycle, bucket, taxicab, cauldron, cardboard box / carton, cassette, storage chest, cocktail shaker, coffee mug, coffeemaker, convertible, crate, electric locomotive, envelope, fire truck, forklift, freight car, garbage truck, goblet, go-kart, golf cart, half-track, hamper, horse-drawn vehicle, jeep, rickshaw, ladle, limousine, messenger bag, mailbox, measuring cup, milk can, minivan, mixing bowl, mobile home, ford model t, moped, mortar and pestle, vespa, mountain bike, moving van, bullock cart, product packet / packaging, railroad car, pencil case, Petri dish, pickup truck, piggy bank, pill bottle, drink pitcher, plastic bag, police van, soda bottle, plant pot, purse, race car, rain barrel, recreational vehicle, safe, salt shaker, shopping basket, shopping cart, sleeping bag, snowmobile, snowplow, soap dispenser, soup bowl, sports car, steam locomotive, tram, tank, teapot, thimble, tow truck, tractor, semi-trailer truck, tray, tricycle, hot tub, unicycle, vase, wallet, sink, water bottle, water jug, water tower, whiskey jug, wine bottle, wooden spoon

\underline{Device:} abacus, accordion, acoustic guitar, analog clock, assault rifle, banjo, barometer, bassoon, binoculars, hunting bow, buckle, candle, cannon, car mirror, carousel, car wheel, automated teller machine, cello, chainsaw, bell or wind chime, combination lock, computer keyboard, cornet, construction crane, desktop computer, digital clock, digital watch, disc brake, drum, electric fan, electric guitar, flute, French horn, gas pump, gong, grand piano, guillotine, hair clip, hand-held computer, hard disk drive, harmonica, harp, combine harvester, hook, hourglass, carved pumpkin, joystick, knot, laptop computer, lighter, music speaker, loupe magnifying glass, magnetic compass, maraca, marimba, maypole, microphone, missile, computer mouse, mousetrap, muzzle, metal nail, neck brace, notebook computer, oboe, ocarina, odometer, oil filter, pipe organ, oxygen mask, paddle wheel, padlock, paintbrush, pan flute, parking meter, plectrum, pier, pinwheel, potter's wheel, power drill, printer, projector, hockey puck, radiator, radio telescope, fishing casting reel, remote control, revolver, rifle, ruler measuring stick, safety pin, saxophone, weighing scale, CRT monitor, screw, ski, slide rule, slot machine, snorkel, solar thermal collector, space heater, spider web, spotlight, steel drum, stethoscope, stopwatch, stove, strainer, sundial, sunglasses, swing, electrical switch, syringe, threshing machine, torch, tripod, trombone, typewriter keyboard, upright piano, vending machine, violin, wall clock, whistle, airplane wing, website

\underline{Instrumentality:} aircraft carrier, airliner, airship, balance beam, balloon, ballpoint pen, barbell, barber chair, baseball, basketball, bassinet, bobsleigh, bookcase, broom, high-speed train, canoe, can opener, tool kit, cassette player, catamaran, CD player, mobile phone, chain, chiffonier, china cabinet, cleaver, container ship, corkscrew, cradle, infant bed, Crock Pot, croquet ball, crutch, desk, rotary dial telephone, dining table, dog sled, drilling rig, drumstick, dumbbell, entertainment center, face powder, filing cabinet, fireboat, flagpole, folding chair, fountain pen, four-poster bed, frying pan, golf ball, gondola, hair spray, hammer, hatchet, gymnastic horizontal bar, iPod, jigsaw puzzle, lawn mower, letter opener, lifeboat, ocean liner, lipstick, lotion, matchstick, maze, medicine cabinet, minibus, modem, monitor, oscilloscope, paddle, parachute, parallel bars, park bench, payphone, pencil sharpener, perfume, photocopier, ping-pong ball, pirate ship, block plane, farm plow, plunger, Polaroid camera, pole, pool table, prayer rug, punching bag, quill, racket, radio, reflex camera, rocking chair, eraser, rugby ball, school bus, schooner, screwdriver, shovel, shower curtain, soccer ball, keyboard space bar, space shuttle, spatula, motorboat, spindle, stretcher, couch, submarine, sunscreen, mop, table lamp, tape player, television, tennis ball, front curtain, throne, toilet seat, trimaran, trolleybus, volleyball, wardrobe, military aircraft, wok, shipwreck, sailboat, comic book, crossword

\underline{Invertebrate:} trilobite, harvestman, scorpion, yellow garden spider, barn spider, European garden spider, southern black widow, tarantula, wolf spider, tick, centipede, jellyfish, sea anemone, brain coral, flatworm, nematode, conch, snail, slug, sea slug, chiton, chambered nautilus, Dungeness crab, rock crab, fiddler crab, red king crab, American lobster, spiny lobster, crayfish, hermit crab, isopod, tiger beetle, ladybug, ground beetle, longhorn beetle, leaf beetle, dung beetle, rhinoceros beetle, weevil, fly, bee, ant, grasshopper, cricket insect, stick insect, cockroach, praying mantis, cicada, leafhopper, lacewing, dragonfly, damselfly, red admiral butterfly, ringlet butterfly, monarch butterfly, small white butterfly, sulphur butterfly, gossamer-winged butterfly, starfish, sea urchin, sea cucumber

\underline{Mammal:} tusker, echidna, platypus, wallaby, koala, wombat, grey whale, killer whale, dugong, sea lion, Chihuahua, Japanese Chin, Maltese, Pekingese, Shih Tzu, King Charles Spaniel, Papillon, toy terrier, Rhodesian Ridgeback, Afghan Hound, Basset Hound, Beagle, Bloodhound, Bluetick Coonhound, Black and Tan Coonhound, Treeing Walker Coonhound, English foxhound, Redbone Coonhound, borzoi, Irish Wolfhound, Italian Greyhound, Whippet, Ibizan Hound, Norwegian Elkhound, Otterhound, Saluki, Scottish Deerhound, Weimaraner, Staffordshire Bull Terrier, American Staffordshire Terrier, Bedlington Terrier, Border Terrier, Kerry Blue Terrier, Irish Terrier, Norfolk Terrier, Norwich Terrier, Yorkshire Terrier, Wire Fox Terrier, Lakeland Terrier, Sealyham Terrier, Airedale Terrier, Cairn Terrier, Australian Terrier, Dandie Dinmont Terrier, Boston Terrier, Miniature Schnauzer, Giant Schnauzer, Standard Schnauzer, Scottish Terrier, Tibetan Terrier, Australian Silky Terrier, Soft-coated Wheaten Terrier, West Highland White Terrier, Lhasa Apso, Flat-Coated Retriever, Curly-coated Retriever, Golden Retriever, Labrador Retriever, Chesapeake Bay Retriever, German Shorthaired Pointer, Vizsla, English Setter, Irish Setter, Gordon Setter, Brittany dog, Clumber Spaniel, English Springer Spaniel, Welsh Springer Spaniel, Cocker Spaniel, Sussex Spaniel, Irish Water Spaniel, Kuvasz, Schipperke, Groenendael dog, Malinois, Briard, Australian Kelpie, Komondor, Old English Sheepdog, Shetland Sheepdog, collie, Border Collie, Bouvier des Flandres dog, Rottweiler, German Shepherd Dog, Dobermann, Miniature Pinscher, Greater Swiss Mountain Dog, Bernese Mountain Dog, Appenzeller Sennenhund, Entlebucher Sennenhund, Boxer, Bullmastiff, Tibetan Mastiff, French Bulldog, Great Dane, St. Bernard, husky, Alaskan Malamute, Siberian Husky, Dalmatian, Affenpinscher, Basenji, pug, Leonberger, Newfoundland dog, Great Pyrenees dog, Samoyed, Pomeranian, Chow Chow, Keeshond, brussels griffon, Pembroke Welsh Corgi, Cardigan Welsh Corgi, Toy Poodle, Miniature Poodle, Standard Poodle, Mexican hairless dog (xoloitzcuintli), grey wolf, Alaskan tundra wolf, red wolf or maned wolf, coyote, dingo, dhole, African wild dog, hyena, red fox, kit fox, Arctic fox, grey fox, tabby cat, tiger cat, Persian cat, Siamese cat, Egyptian Mau, cougar, lynx, leopard, snow leopard, jaguar, lion, tiger, cheetah, brown bear, American black bear, polar bear, sloth bear, mongoose, meerkat, cottontail rabbit, hare, Angora rabbit, hamster, porcupine, fox squirrel, marmot, beaver, guinea pig, common sorrel horse, zebra, pig, wild boar, warthog, hippopotamus, ox, water buffalo, bison, ram (adult male sheep), bighorn sheep, Alpine ibex, hartebeest, impala (antelope), gazelle, arabian camel, llama, weasel, mink, European polecat, black-footed ferret, otter, skunk, badger, armadillo, three-toed sloth, orangutan, gorilla, chimpanzee, gibbon, siamang, guenon, patas monkey, baboon, macaque, langur, black-and-white colobus, proboscis monkey, marmoset, white-headed capuchin, howler monkey, titi monkey, Geoffroy's spider monkey, common squirrel monkey, ring-tailed lemur, indri, Asian elephant, African bush elephant, red panda, giant panda

\underline{Vertebrate:} tench, goldfish, great white shark, tiger shark, hammerhead shark, electric ray, stingray, fire salamander, smooth newt, newt, spotted salamander, axolotl, American bullfrog, tree frog, tailed frog, loggerhead sea turtle, leatherback sea turtle, mud turtle, terrapin, box turtle, banded gecko, green iguana, Carolina anole, desert grassland whiptail lizard, agama, frilled-necked lizard, alligator lizard, Gila monster, European green lizard, chameleon, Komodo dragon, Nile crocodile, American alligator, triceratops, worm snake, ring-necked snake, eastern hog-nosed snake, smooth green snake, kingsnake, garter snake, water snake, vine snake, night snake, boa constrictor, African rock python, Indian cobra, green mamba, sea snake, Saharan horned viper, eastern diamondback rattlesnake, sidewinder rattlesnake, snoek fish, eel, silver salmon, rock beauty fish, clownfish, sturgeon, gar fish, lionfish, pufferfish

\subsection{Multiple Choice Question Generation}

To extract conceptual supervision from the language model, we prompt it to generate multiple-choice questions that help distinguish between visual classes. These questions serve as a bridge between semantic knowledge encoded in the LLM and the class-level structure of the visual dataset. Each question is designed to capture a visual or contextually related attribute that differentiates one class from another, and the resulting class-wise logits from the LLM are used as distillation targets. Below, we provide the exact prompt used to generate 50 questions with five answer options per question.

\begin{verbatim}
 Your task:
1. Generate 50 questions for 
    distinguishing between the classes in 
    a dataset with the requirements below.
2. Each question should be centered around
    visual concepts while 
    slight deviation is acceptable. 
    An example of a deviation would 
    be about the environment.
3. Each question should have 5 answer 
    options and each class can 
    only have one correct answer option. 
    It’s best to maximize 
    the number classes that each pick 
    a different answer option.
4. Each question should contain 
    “the class” in the question.
5. Questions should maximize the 
    separation between classes like a 
    decision tree maximizing entropy.
6. Use your understanding of all of 
    the classes and their visual 
    differences to create these questions. 
7. Only output ALL of the questions 
    and answer options.
8. Do not repeat questions.
9. Do not write code.
10. Do not include class names in the 
    answer options.

The classes:
<classes>

Output format:
- For each question, use the specific
    format:
	[Question]
	1. [Option 1]
…
- Do not add additional commentary.
- Do not include the square brackets 
in the answer.
- Do not number the questions.   
\end{verbatim}

\begin{table*}[h!]
\centering
\begin{tabular}{lcccccccc}
\toprule
\textbf{Student} & \textbf{Dataset} & \textbf{Method} & \textbf{-1} & \textbf{-6} & \textbf{-12} & \textbf{-18} & \textbf{-24} & \textbf{-30} \\
\midrule
\multirow{12}{*}{RN18} & \multirow{2}{*}{CUB} & FitNet & 63.51 & 63.75 & 62.95 & 63.64 & 63.73 & 63.51 \\
& & CRD & 69.13 & 68.81 & 69.47 & 69.32 & 68.59 & 66.12 \\
\cmidrule(lr){2-9}
& \multirow{2}{*}{Caltech} & FitNet & 78.52 & 78.93 & 78.87 & 79.11 & 78.97 & 79.17 \\
& & CRD & 80.15 & 80.19 & 80.83 & 80.20 & 80.44 & 79.97 \\
\cmidrule(lr){2-9}
& \multirow{2}{*}{Flowers} & FitNet & 76.13 & 76.37 & 76.47 & 76.55 & 76.14 & 75.91 \\
& & CRD & 80.75 & 81.03 & 80.93 & 80.64 & 76.24 & 76.08 \\
\cmidrule(lr){2-9}
& \multirow{2}{*}{Aircraft} & FitNet & 80.19 & 80.21 & 79.95 & 79.80 & 79.89 & 80.15 \\
& & CRD & 79.42 & 79.72 & 80.81 & 80.98 & 81.16 & 79.50 \\
\cmidrule(lr){2-9}
& \multirow{2}{*}{Pets} & FitNet & 77.02 & 76.76 & 76.91 & 76.82 & 77.70 & 77.08 \\
& & CRD & 81.80 & 81.61 & 81.80 & 81.40 & 81.50 & 79.36 \\
\cmidrule(lr){2-9}
& \multirow{2}{*}{Cars} & FitNet & 85.84 & 86.04 & 85.90 & 86.06 & 86.08 & 86.20 \\
& & CRD & 86.78 & 86.71 & 86.86 & 87.39 & 86.87 & 86.23 \\
\midrule
\multirow{12}{*}{MNV2} & \multirow{2}{*}{CUB} & FitNet & 69.70 & 70.00 & 70.00 & 69.84 & 69.56 & 70.03 \\
& & CRD & 71.90 & 71.35 & 71.96 & 71.72 & 70.86 & 70.06 \\
\cmidrule(lr){2-9}
& \multirow{2}{*}{Caltech} & FitNet & 81.91 & 81.95 & 82.27 & 81.70 & 81.73 & 81.30 \\
& & CRD & 78.87 & 79.03 & 79.00 & 79.64 & 79.01 & 80.28 \\
\cmidrule(lr){2-9}
& \multirow{2}{*}{Flowers} & FitNet & 83.52 & 83.52 & 83.46 & 83.10 & 84.08 & 83.51 \\
& & CRD & 85.65 & 85.69 & 85.21 & 86.25 & 83.43 & 83.29 \\
\cmidrule(lr){2-9}
& \multirow{2}{*}{Aircraft} & FitNet & 85.05 & 84.72 & 85.46 & 85.55 & 85.14 & 84.86 \\
& & CRD & 82.56 & 82.69 & 82.31 & 82.78 & 82.91 & 82.47 \\
\cmidrule(lr){2-9}
& \multirow{2}{*}{Pets} & FitNet & 80.38 & 80.88 & 80.57 & 80.77 & 78.12 & 77.96 \\
& & CRD & 77.56 & 82.89 & 83.83 & 83.50 & 81.86 & 80.38 \\
\cmidrule(lr){2-9}
& \multirow{2}{*}{Cars} & FitNet & 86.65 & 86.55 & 86.77 & 87.05 & 87.02 & 87.01 \\
& & CRD & 86.24 & 86.02 & 86.59 & 86.73 & 86.58 & 86.51  \\
\midrule
\multirow{12}{*}{SNV2} & \multirow{2}{*}{CUB} & FitNet & 65.42 & 65.54 & 64.82 & 65.80 & 65.63 & 65.20 \\
& & CRD & 67.53 & 68.00 & 68.17 & 68.16 & 67.05 & 65.69 \\
\cmidrule(lr){2-9}
& \multirow{2}{*}{Caltech} & FitNet & 78.88 & 79.03 & 79.39 & 78.93 & 79.30 & 78.89 \\
& & CRD & 78.50 & 78.15 & 77.95 & 78.51 & 78.36 & 78.72 \\
\cmidrule(lr){2-9}
& \multirow{2}{*}{Flowers} & FitNet & 80.52 & 80.84 & 81.22 & 80.42 & 80.82 & 80.76 \\
& & CRD & 82.57 & 82.72 & 82.44 & 83.12 & 80.30 & 80.78 \\
\cmidrule(lr){2-9}
& \multirow{2}{*}{Aircraft} & FitNet & 80.13 & 80.63 & 80.61 & 80.44 & 80.67 & 80.35 \\
& & CRD & 77.82 & 78.40 & 78.78 & 78.83 & 77.85 & 78.52 \\
\cmidrule(lr){2-9}
& \multirow{2}{*}{Pets} & FitNet & 77.56 & 78.30 & 77.78 & 77.68 & 77.74 & 77.87 \\
& & CRD & 80.69 & 80.43 & 81.28 & 80.49 & 80.41 & 77.84 \\
\cmidrule(lr){2-9}
& \multirow{2}{*}{Cars} & FitNet & 85.37 & 85.13 & 85.04 & 85.78 & 84.88 & 84.72 \\
& & CRD & 84.26 & 84.26 & 84.83 & 85.34 & 84.45 & 84.36 \\
\bottomrule
\end{tabular}

\caption{Top-1 accuracy (\%) for student models trained with feature-based distillation from different LLM layers of LLaVA-1.5. We compare FitNet and CRD across six fine-grained classification datasets, using three student architectures: ResNet-18 (RN18), MobileNetV2 (MNV2), and ShuffleNetV2 (SNV2). Each column corresponds to a different LLaVA transformer layer, with “-1” indicating the closest to output layer and “-30” the closest to input layer. Mid-to-late layers often yield the best results, indicating that semantically rich supervision emerges progressively within the LLM.}
\label{tab:layers}
\end{table*}

\section{Ablation Study on Feature-Based Distillation from a Multimodal Teacher}
\label{sec:ablation_llava}

To establish stronger feature-based LLM baselines for comparison, we adapt two representative knowledge distillation methods—FitNet~\cite{romero2014fitnets} and Contrastive Representation Distillation (CRD)~\cite{tian2019contrastive}—to use LLaVA~\cite{liu2023visual} as the teacher. These serve as key baselines in our main evaluation (Section 4.2). Since traditional feature-based KD methods rely on matching internal activations, we extract features from various layers of LLaVA and assess their impact on student performance. We find that distillation performance varies substantially by layer, motivating a layer-wise ablation study to fairly configure each baseline.

\paragraph{Experimental Setup}
We use the standard LLaVA-1.5 model and follow its prompting format, where the image is prepended to the language prompt using a special token (e.g., \texttt{<image>}). The full prompt is structured as a user-assistant exchange:
\begin{quote}
\texttt{USER: <image> Is there a <class> in this image? \\ ASSISTANT:}
\end{quote}
We feed this prompt into LLaVA and extract the embedding of the final token at each transformer layer of the LLM. This token embedding reflects the fused multimodal representation at various levels of abstraction. We then use this as the distillation target for training a vision-only student model. Distillation is applied using either FitNet (with $\ell_2$ regression) or CRD (with contrastive learning), and we vary the teacher layer from which the token embedding is extracted.

\paragraph{Feature Layer Selection for Feature-based Distillation}
Table~\ref{tab:layers} reports top-1 accuracy on the same main six classification datasets using ResNet-18, MobileNetV2, and ShuffleNetV2 as student architectures. Each column corresponds to a different LLaVA LLM transformer layer, with “-1” representing the final layer and “-30” the earliest layer. We observe that mid-to-late layers (e.g., -12 to -18) tend to produce stronger supervision signals, suggesting that class-level semantic structure becomes more explicit in deeper LLM layers. Neither method is consistently outperforming the other, however, in most cases, CRD does do better, which reflects the strength of contrastive alignment in high-dimensional spaces. Overall, because of their inconsistent performance, we illustrate that \LaViD{} is still superior in harnessing an LLM for vision distillation, even without the vision modality.

\end{document}